\documentclass[10pt,twocolumn,letterpaper]{article}

\usepackage{iccv}
\usepackage{times}
\usepackage{epsfig}
\usepackage{graphicx}
\usepackage{amsmath}
\usepackage{amssymb}
\usepackage{amsthm}
\usepackage{xcolor}
\usepackage{multirow}
\usepackage{lipsum}

\newcommand{\bx}{\mathbf{x}}
\newcommand{\by}{\mathbf{y}}

\newcommand{\dom}{\textbf{dom}}
\newcommand{\deriv}{\mathrm{d}}

\newcommand{\tb}[1]{\textbf{#1}}

\newtheorem{definition}{Definition}
\newtheorem{proposition}{Proposition}
\newtheorem*{proposition*}{Proposition}
\newtheorem*{remark}{Remark}

\usepackage[pagebackref=true,breaklinks=true,letterpaper=true,colorlinks,bookmarks=false]{hyperref}

\iccvfinalcopy 

\ificcvfinal\fi
\begin{document}

\title{Deformable Filter Convolution for Point Cloud Reasoning}

\author{Yuwen Xiong\thanks{Equal contribution} $^{\ 1,2}$  ,\ Mengye Ren$^{* 1, 2}$, Renjie Liao$^{1, 2}$, Kelvin Wong$^{1, 2}$, Raquel Urtasun$^{1, 2}$\\
$^{1}$Uber Advanced Technologies Group\\
$^{2}$University of Toronto\\
{
\tt\small \{yuwen,mren3,rjliao,kelvin.wong,urtasun\}@uber.com}
}

\maketitle

\begin{abstract}
Point clouds are the native output of many real-world 3D sensors. To borrow the success of 2D
convolutional network architectures, a majority of popular 3D perception models voxelize the
points, which can result in a loss of local geometric details that cannot be recovered. In this
paper, we propose a novel learnable convolution layer for processing 3D point cloud data directly.
Instead of discretizing points into fixed voxels, we deform our learnable 3D filters to match with
the point cloud shape. We propose to combine voxelized backbone networks with our deformable
filter layer at 1) the network input stream and 2) the output prediction layers to enhance point
level reasoning. We obtain state-of-the-art results on LiDAR semantic segmentation and producing a
significant gain in performance on LiDAR object detection.

\end{abstract}

\section{Introduction}
3D perception is one of the key components of  real-world robotic systems. These robots are typically equipped with 3D sensors such as LiDAR and RGBD
cameras which produce outputs in the form of point clouds. These point clouds correspond to a set of vectors of location coordinates and
associated features. Driven by the success of deep learning on 2D images, there has been a fresh
wave of deep network architectures proposed to tackle this new challenge: unlike image grids, point
clouds are sampled sparsely and non-uniformly with continuous spatial coordinates, and they are
equivalent up to permutations.

A well-studied approach is to discretize the points into voxels \cite{voxnet}. As 3D convolution can
be inefficient in terms of both computation and storage, various approximations have been proposed
\cite{octnet,ocnn} -- recent work also finds 2D convolutions almost as competitive but with a much
improved efficiency \cite{pixor,effconv,sbnet}. Despite great strides being made by leveraging existing
2D CNN backbone networks on voxelized inputs, current approaches face the dilemma of either clashing
with a loss of local geometric details or struggling with sensor noise and a smaller field of view.

To complement the weaknesses of voxelized networks, new network architectures have been proposed to
process points directly. While simple average and max-pooling operations can preserve permutation
invariance \cite{pointnet,pointnet2,deepsets}, this approach lacks interpretability on how spatial features
are aggregated. Various attempts to define a general learnable continuous convolution layer have
been made, by hammering a learned multi-layer network to predict the filter weights or features
\cite{pccn,contfuse,pointcnn,mcconv}, or restricting to a simpler family of functions \cite{spidercnn,schnet}.
While the former may lack robustness and require extra supervision for regions with sparse
neighborhoods, the latter could be limited by less powerful feature representations.

This paper advocates a simple idea: we learn 3D cuboid filters just like the ones in a voxelized
network. However, when performing the convolution operation, instead of discretizing the points into a 
voxel grid, we deform the 3D filter towards the point clouds. We propose two ways to integrate our
proposed convolution operator into popular backbones, by 1) enriching feature representation from an
additional point-wise input stream, and by 2) smoothing out point-wise predictions within local
neighborhoods. Composing our deformable filter convolution layers with voxelized networks results in
a significant gain in performance on semantic segmentation and object detection tasks, comparing to
voxel only networks and previous attempts at fusing point-wise features. Moreover, our proposed
joint network achieves state-of-the-art performance on the TOR4D large-scale LiDAR semantic
segmentation benchmark.


\section{Related work}
Previous work on processing 3D data can be roughly categorized into multi-viewpoints, voxelization,
and point-based representations. 
Inspired by an agent-centric 2D view of the world, multi-view
representations \cite{mvcnn,voxmv,mv3d} treat 3D data as snapshots of 2D images taken at different
view points. Front view representations \cite{velofcn}, which considers depth as an additional
channel in the input, can leverage  2D image network architectures. However, these
approaches bear a significant loss of 3D information, and are unable to reason about 3D rigid
transformations. To address this issue, voxelization-based representations instead process the
occupancy grids using 3D convolution \cite{voxnet}. As pure 3D convolution suffers from
computation and storage inefficiency, OctNet \cite{octnet} and O-CNN \cite{ocnn} use OctTree to
efficiently compute voxel convolutions.

While voxels are intuitive 3D counterparts to 2D images, real world sensors such as LiDAR and
depth camera instead produce point clouds as their native output. A popular approach is to
discretize points with point statistics in each voxel of a grid and learn a 2D convolutional
neural net (CNN) using a bird's eye view representation \cite{mv3d,faf,pixor,sbnet,hdnet}. As such
procedure can potentially result in loss of details, especially when the points are sampled
non-uniformly, deep network architectures have hence been designed to directly handle point data
as inputs. Qi \etal \cite{pointnet} propose PointNet which applies a fully connected network on
each point individually and a permutation-invariant max-pooling operation to aggregate global
information. To leverage local neighborhood and hierarchical information passing, PointNet++
\cite{pointnet2} adds grouping and sampling layers to perform stagewise aggregation, which is
similar to pooling layers in regular CNNs. \cite{fpointnet,pointcnn} demonstrate the effectiveness
of the PointNet-based architecture on 3D object detectors. Inspired by the SIFT feature extractor
\cite{sift}, Jiang
\etal propose PointSIFT \cite{pointsift}, which uses local octant directional vectors as feature
extraction layers showing good results on point cloud segmentation tasks.

Grouping local neighborhood of points and aggregating information using permutation invariant
operators, as done in PointNet++~\cite{pointnet2}, are special cases of graph neural networks
(GNNs) \cite{spectualnet,localspectralcnn,gcn,ggnn}, where node interactions are modeled using a
neural network. Point clouds can be treated as a sparse graph where edges denote two points which
are close. 3D-GNN \cite{3dgnn} uses a GNN to approximate messaging passing in point clouds. ECC
\cite{ecc} and EdgeConv \cite{dynamicgcnn} propose to generate the edge weights through a neural
network. KD-Net \cite{kdnet} recursively processes hierarchical graph structures through a
KD-Tree. \cite{kcnet} uses learnable kernel anchor points to smooth out local neighborhoods.

Another line of work views point clouds as discrete samples of a continuous function in space.
Various parameterizations of learnable continuous filter functions have thus been proposed. Wang
\etal \cite{pccn} and Hermosilla \etal \cite{mcconv} propose to use a multi-layer perceptron (MLP)
to represent the convolution filter function. The MLP takes in an offset vector towards the center
of the local neighborhood, and outputs the value of the filter function at that location. 
ContFuse~\cite{contfuse} is a memory efficient successor of \cite{pccn} as it directly predicts the output features through an MLP.
Other
families of learnable filter functions have also been studied, e.g. radial basis function (RBF)
\cite{schnet} and polynomial function kernels \cite{spidercnn}. To prevent the function value from
growing unbounded at a large distance, a step function is applied in \cite{spidercnn} to make sure
the filter function is zero outside a certain radius. In contrast, instead of predicted by a
parametric function \cite{contfuse,pccn,mcconv,schnet,spidercnn}, our 3D filters have learnable weights at
well defined 3D positions, which are potentially more robust and sample efficient.

Motivated by 3D convolution with grid structured filters, Atzmon \etal \cite{extop} propose
``extension'' and ``restriction'' operators. First, the extension operator interpolates point
features onto a grid structure; then 3D convolution is applied on the grids; finally, the
restriction operator projects convolved features back to point locations. Similarly, SPLATNet
\cite{splatnet} extends points onto lattice grids, and PointwiseCNN \cite{pointwisecnn}
discretizes points into filter bins. These extension operators could potentially suffer from the
loss of local geometric information due to discretization. The design of our deformable filter
convolution also takes inspiration from deformable convolution~\cite{deformable}, extending model
capacity with continuous spatial reasoning. Despite the main difference of applying on 2D images
vs. 3D point clouds, our proposed operator does not resample point features, as was done in
\cite{deformable}, but interpolates 3D filters at point locations. This particular design
addresses the potential discretization issue mentioned above \cite{extop,splatnet}. In concurrent work  KPConv~\cite{thomas2019KPConv}  also tries to deform filters similarly to our approach. 





\begin{figure*}[t]
\centering
\includegraphics[width=0.96\textwidth,trim={0 0 3cm 0},clip]{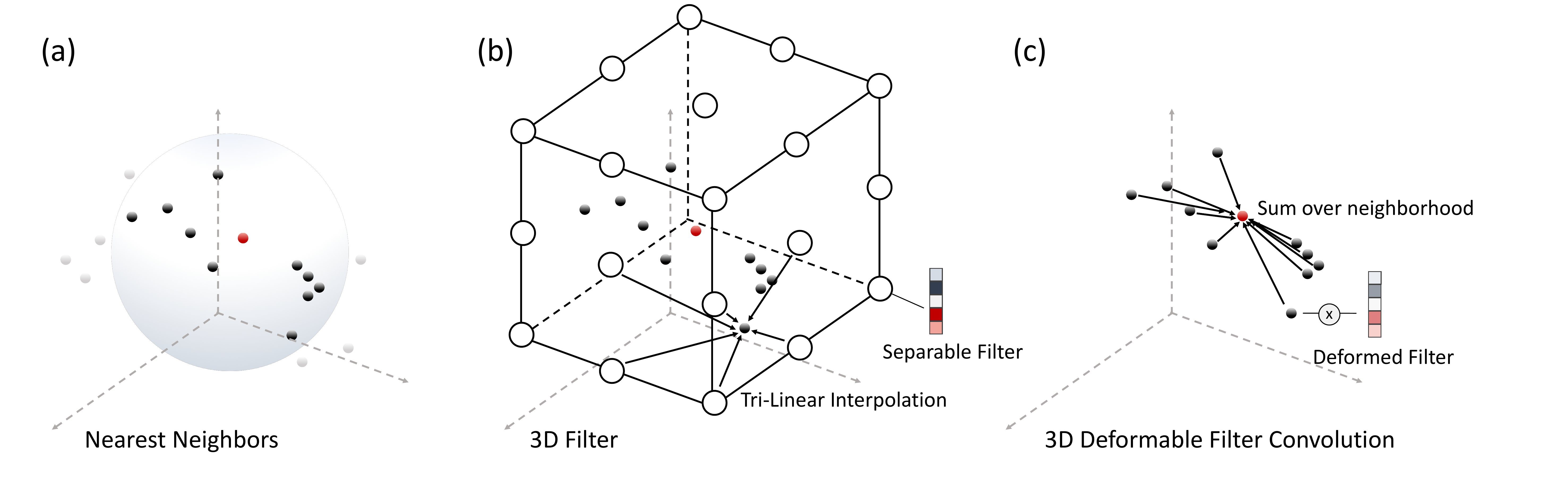}
\caption{Overview of the proposed deformable filter convolution on 3D point cloud data. \textbf{(a):}
At a given centroid point (\textcolor{red}{red}), a neighborhood of points are searched.
\textbf{(b):} Each point in the neighborhood retrieves its own filter by performing tri-linear
interpolation over the $3\times3\times3$ filter anchors. \textbf{(c):} The filter is multiplied with
the point features, and summed towards the centroid point. To speed up computation, we use separable
filters, and the spatial filter is of shape $3\times3\times3\times D$, where $D$ is the feature
dimension.}
\label{fig:overview}
\end{figure*}

\begin{figure}[t]
\centering
\includegraphics[width=0.98\columnwidth,trim={0 10cm 10cm 0},clip]{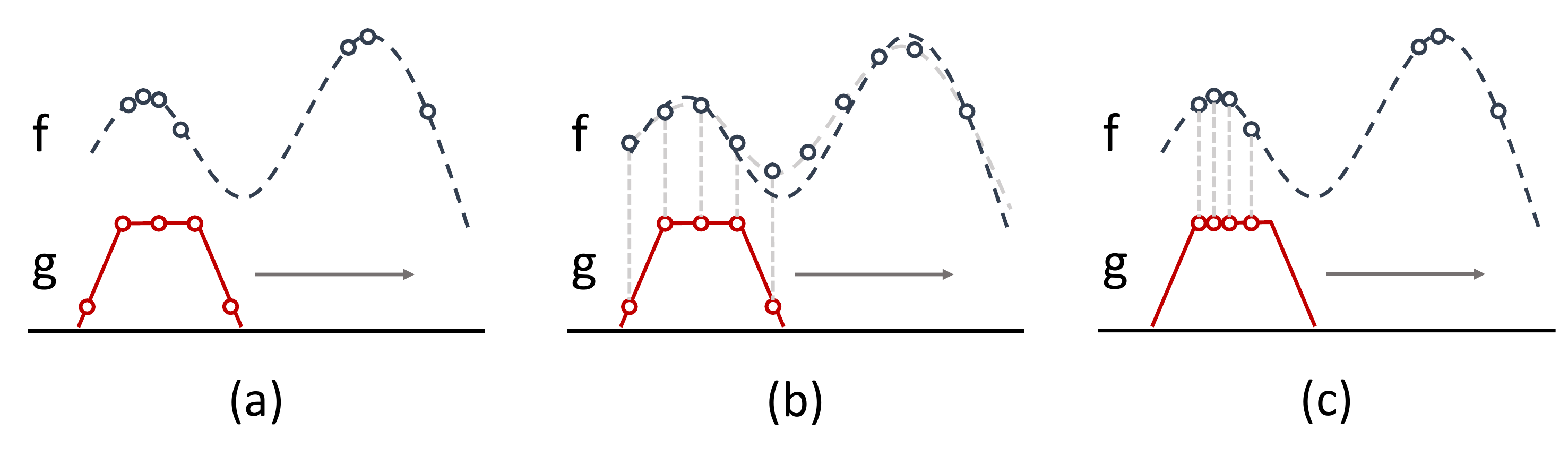}
\caption{Our proposed deformable filter operator vs. extension operator \cite{extop}. \textbf{(a):}
Convolution of point cloud signal $f$ (under non-uniform sampling) and filter $g$. \textbf{(b):}
Extension operator \cite{extop} resamples signal at grid locations, which can potentially lose local
geometric details. \textbf{(c):} Our deformable filter convolution, which resamples the filter at
positions of the point clouds, preserving local geometric details.}
\label{fig:1dconv}
\end{figure}

\section{Deformable Filter Convolution}
In this section, we first define our deformable filter convolution operator in the context of
spatially continuous functions with discrete samples.  We then show equivariance properties
of the proposed operator.

\subsection{Deformable Filters}

Let's consider a spatially continuous signal $f: \mathbb{R}^3 \mapsto \mathbb{R}^D$ to represent
the features from a 3D world. $f$ takes in a 3-D coordinate and returns the features at that
location. Point clouds can be viewed as discrete samples of this continuous function. Convolution
in the continuous domain can be written as:
\begin{align}
h(\by) = (f * g)(\by) \equiv \int_{-\infty}^{\infty} f(\bx) \cdot g(\by - \bx) \deriv \bx,
\end{align}
where $f$ is the original signal, and $g: \mathbb{R}^3 \mapsto \mathbb{R}^D$ is the 3D filter. 
Computing this integral is, however, difficult in most cases. To make this computation tractable,
we discretize the signal and approximate the integral using a Monte Carlo estimate of the local
neighborhood, which is valid as long as the support of $g$ is a subset of the neighborhood:
\begin{align}
h(\by) \approx \sum_{\bx \in {\cal{N}}(\by)} \mu(\bx) f(\bx) \cdot g(\by - \bx),
\end{align}
where ${\cal N}(\by) = \{\bx: \lVert \bx - \by \rVert \le r\}$ is the set of points in the
neighborhood of $\by$, and $\mu$ is the measure of the volume covered by each neighboring point.
For simplicity, we further assume that $\mu$ is some constant based on the  approximation that
points are uniformly sampled in local neighborhoods. Without loss of generality, we assume $\mu=1$
since the actual value can be merged with the learned filter $g$.

Previous work proposed to represent the filter $g(\by - \bx)$ using an MLP \cite{pccn,mcconv} or a
polynomial function \cite{spidercnn}. The potential issues with these continuous representations are
1) the filter kernel can be highly non-linear, and 2) depending on the point cloud distribution, the
actual filter being used can vary significantly.

Different from previous approaches, we only parameterize the filter at discretized anchors $X'$ that
are coherent with the 3D grid structure.
In contrast to \cite{extop}, which ``voxelizes'' the points into a grid structure and
then projects them back to the original locations, we ``deform'' the standard 3D filter from the anchors
towards the points $(\by - \bx)$. This leads to better preservation of local geometric information
compared to feature voxelization as shown in Figure~\ref{fig:1dconv} for the 1D case. To deform 3D
filters, we use an interpolation kernel $k(\cdot, \cdot)$,
\begin{align}
\hat{g}(\by - \bx) &\approx \sum_{\bx' \in X'} k(\bx', \by - \bx) g(\bx').
\end{align}
In practice, we choose to use a tri-linear interpolation kernel 
\begin{align}
k(\bx, \bx') = 
\prod_{d=1}^3 \max\left(1 - \frac{\lvert \bx_d - \bx'_d \rvert}{a_d}, 0\right),
\label{eq:trilinear}
\end{align}
where $a_d$ is the filter grid unit length on dimension $d$. This interpolation scheme is
continuous everywhere and naturally decays to zero when a point falls far away from the anchors,
without using a manually designed step function, as was done in \cite{spidercnn}. Tri-linear
interpolation is easy to implement and unlike the Gaussian kernel, it does not have the gradient
vanishing problem. In summary, our 3D deformable filter convolution operator can be written as:
\begin{align}
h(\by) = 
\sum_{\bx \in {\cal{N}}(\by)} f(\bx) \cdot \left[ \sum_{\bx' \in X'} k(\bx', \by - \bx) g(\bx') \right].
\end{align}

\paragraph{Separable convolution:} To make our convolution operator more efficient, in practice we
consider using {\it separable convolution} \cite{separable}, where we approximate $g(\bx) \approx
g_1(\bx) g_2$. $g_1:
\mathbb{R}^3 \mapsto \mathbb{R}$ is a spatial filter, and $g_2 \in
\mathbb{R}^D$ is a $D$-dimensional vector. Figure~\ref{fig:overview} illustrates the overview of applying our deformable filter convolution on 3D point clouds.

\subsection{Analysis}

In this section we analyze various equivariance properties of the proposed convolution operator
on point clouds. First, we show that our operator is translation-equivariant. 
Second, we show that our operator is permutation-equivariant under discretization of continuous signals. 
Equivariance could be more useful than invariance in
certain scenarios as it preserves the transformation information. We start by defining the
equivariance property mathematically.
\begin{definition}[Equivariance]
Let $\mathcal{T}^F_g: F \mapsto F$ be a transformation operator that produces a group action of $g$
in a transformation group $G$ on a function space F. An operator $L: F
\mapsto H$ is said to be $\mathcal{T}_G$-equivariant if $L(\mathcal{T}^F_g(f)) =
\mathcal{T}^H_g(L(f))$ for any $f \in F$, $g \in G$.
\end{definition}

Translation equivariance is a desired property as it is an efficient way of sharing parameters that
produces consistent outputs regardless of the location of the regions of interest. In short, if the
input $f$ is translated by an offset $\Delta \bx$, the effect on the output $h$ is also a
translation of $\Delta \bx$. CNNs have translation equivariance on 2D grid locations, which is one
of the reasons they are successful in the  image domain. Here, we show that our convolution
operator is translation equivariant in a continuous domain with a $d$-dimensional coordinate
system. 
In contrast, popular voxelization based approaches (e.g., \cite{voxnet,mv3d,pixor,extop,splatnet}) are
unfortunately not translation equivariant since the voxel grid is fixed and points can be assigned
to different discretization bins.
\begin{definition}[Translation operator] $\mathcal{T}^F_{\Delta \bx}(\cdot): F \mapsto F$ is a
translation operator on F if $\mathcal{T}_{\Delta
\bx}(f)(\bx)=f(\bx + \Delta\bx)$  for all $f \in F$, $\bx \in \dom(f)$.
\end{definition}
\begin{proposition}[Translation equivariance]
\label{translation}
Let $C_{g}(\cdot): F \mapsto H$ be the deformable filter convolution operator, where
$F = \{f: \mathbb{R}^d \mapsto \mathbb{R}^{D'}\}$ is the set of input functions, $g:
\mathbb{R}^d \mapsto \mathbb{R}^{D'\times D}$ is the convolution filter, and $H = \{h: \mathbb{R}^d
\mapsto
\mathbb{R}^{D}\}$ is the set of output functions. For all $\by \in
\mathbb{R}^{d}$, $C_{g}(\mathcal{T}^F_{\Delta \bx}(f))(\by) = \mathcal{T}^H_{\Delta
\bx}(C_{g}(f))(\by)$.
\end{proposition}
\begin{proof}See Appendix~\ref{sec:proofs}.\end{proof}

\begin{remark}
Note that Proposition~\ref{translation} is generalized to any coordinate system
of the input points. In Cartesian coordinates of 3D space, $d=3$ and translation means the
conventional translation, whereas translation in polar coordinates is equivalent to rotation in
Cartesian coordinates.
\end{remark}

When the inputs are point clouds, the input function is discretized by an input array, where each
entry stores $D'$-dimensional features of the point. The output of the convolution operation is an
array with output $D$-dimensional features of the point. Permutation equivariance ensures that the
ordering of the points in the input does not affect the output. PointNet \cite{pointnet} aggregates
information using a global max-pooling, which is permutation-invariant but not equivariant. As a
result, it cannot aggregate local neighborhood information.

We first define the permutation operator on the set of functions with integer domain. Note that all
arrays can be represented as a function that maps from positive integers to numbers.
\begin{definition}[Permutation operator]
Let $F = \{f: \textbf{dom}(f)=X\subseteq \mathbb{Z}\}$, where $\mathbb{Z}$ is the set of integers.
Given a permutation $s \in \text{Sym}(X)$, $\mathcal{P}^F_s(\cdot): F \mapsto F$ is a permutation
operator if $\mathcal{P}^F_s(f)(i) = f(s(i))$ for all $i \in X$.
\end{definition}

Now consider an array of $M$ input points $p: \mathbb{Z}_M \mapsto \mathbb{R}^O$. The neighborhood
function on this discretized domain is defined as ${\cal{N}}(i) = \{ j: \lVert p(j) - p(i)
\rVert \le r \}$.
\begin{proposition}[Permutation equivariance]
\label{permutation}
Let $\tilde{C}_{g}(\cdot,\cdot): P \times F \mapsto H$ be the deformable filter convolution operator on
discretized inputs of $M$ points, where $P = \{p: \mathbb{Z}_M \mapsto \mathbb{R}^d\}$ is the set of input coordinate arrays, $F = \{f: \mathbb{Z}_M
\mapsto
\mathbb{R}^{D'}\}$ is the set of input feature arrays, $g:
\mathbb{R}^d \mapsto \mathbb{R}^{D'\times D}$ is the convolution filter, and $H = \{h: \mathbb{Z}_M
\mapsto \mathbb{R}^{D}\}$ is the set of output feature arrays. For all $i \in \mathbb{Z}_M$, $s \in \text{Sym}(M)$,
$C_{g}(\mathcal{P}^{P \times F}_s(p,f))(i) = \mathcal{P}^H_s(C_{g}(p,f))(i)$.
\end{proposition}
\begin{proof}See Appendix~\ref{sec:proofs}.\end{proof}

 \vspace{-0.1in}
\section{Experimental Evaluation}
We verify the effectiveness of our proposed operator on a suite of point cloud benchmarks and tasks 
including semantic segmentation, object detection, and object classification.
\subsection{TOR4D point cloud segmentation}
\begin{table*}[t]
\vspace{-0.2in}
\begin{center}
\begin{small}
\begin{tabular}{|c|c|cccccc|}
\hline
Method                      & mIOU       & Vehicle     & Bicyclist  & Pedestrian & Motorcycle & Background  & Road       \\
\hline\hline                                                                                                                
PointNet \cite{pointnet}    & 46.00      & 76.73       & 2.85       & 6.62       & 8.02       & 89.83       & 91.96      \\   
3D-FCN \cite{pccn}          & 57.33      & 86.74       & 22.30      & 38.26      & 17.22      & 86.91       & 92.56      \\
3D-FCN +PCC \cite{pccn}     & 67.53      & 91.83       & 40.23      & 47.74      & 42.91      & 89.27       & 93.18      \\
2D-U-Net \cite{effconv}     & 60.73      & 91.15       & 27.41      & 51.44      & 41.19      & 92.45       & 87.82      \\
3D-U-Net \cite{effconv}     & 67.15      & 91.29       & 43.35      & 43.91      & 45.01      & 92.40       & 86.96      \\
\hline                                                                                                                     
U-Net \cite{unet}           & 73.69      & 91.78       & 52.18      & 60.32      & \tb{51.01} & 91.53       & 95.32      \\
+ContFuse \cite{contfuse}   & 76.93      & 94.81       & 58.72      & \tb{78.53} & 38.01      & \tb{94.94}  & \tb{96.59} \\
+DeformFilter (Ours)        & \tb{79.19} & \tb{94.93}  & \tb{60.86} & 77.96      & 50.16      & 94.71       & 96.50      \\ 
\hline
\end{tabular}
\end{small}
\vspace{0.1in}
\caption{TOR4D point cloud semantic segmentation results on test set}
\label{tab:semseg}
\end{center}
\vspace{-0.15in}
\end{table*}
\begin{table*}[t]
\begin{center}
\begin{small}
\begin{tabular}{|c|ccc|ccc|ccc|}
\hline 
\multirow{2}{*}{Method} & \multicolumn{3}{|c|}{Car} & \multicolumn{3}{|c|}{Pedestrian} & \multicolumn{3}{|c|}{Cyclist} \\
                          & Easy       & Moderate  & Hard       & Easy       &  Moderate  &  Hard      & Easy       & Moderate   & Hard           \\
\hline\hline                                                                                                                                        
HDNet \cite{hdnet}        & 90.67      & 86.99     & 85.32      & 73.23      &  69.88     &  66.41     & 78.41      & 64.14      & 60.80          \\
+PCC \cite{pccn}          & 92.67      &\tb{87.46} & 84.71      & 73.49      &  70.43     &  66.19     & 74.18      & 59.84      & 57.17          \\
+DeformFilter (Ours)      & \tb{92.71} & 87.30	   & \tb{85.48} & \tb{75.24} & \tb{71.35} & \tb{67.31} & \tb{79.57} & \tb{65.41} & \tb{62.86}     \\
\hline
\end{tabular}
\end{small}
\vspace{0.1in}
\caption{KITTI BEV object detection results on val set}
\label{tab:kitti_val}
\end{center}
\end{table*}
First, we verify the usefulness of the proposed deformable filter convolution on the task of LiDAR
semantic segmentation. We report results on the TOR4D dataset \cite{faf,pccn}, which consists of
1,239,706 frames in training, 123,975 frames in validation and 123,475 frames in test set. 
The dataset contains 7 object
classes: ``vehicle'', ``bicyclist'', ``pedestrian'', ``motorcycle'', ``background'', ``animal'', and
``road''. We omit the ``animal'' class for evaluation due to the small number of examples. 

\begin{figure}[t]
\centering
\includegraphics[width=\columnwidth]{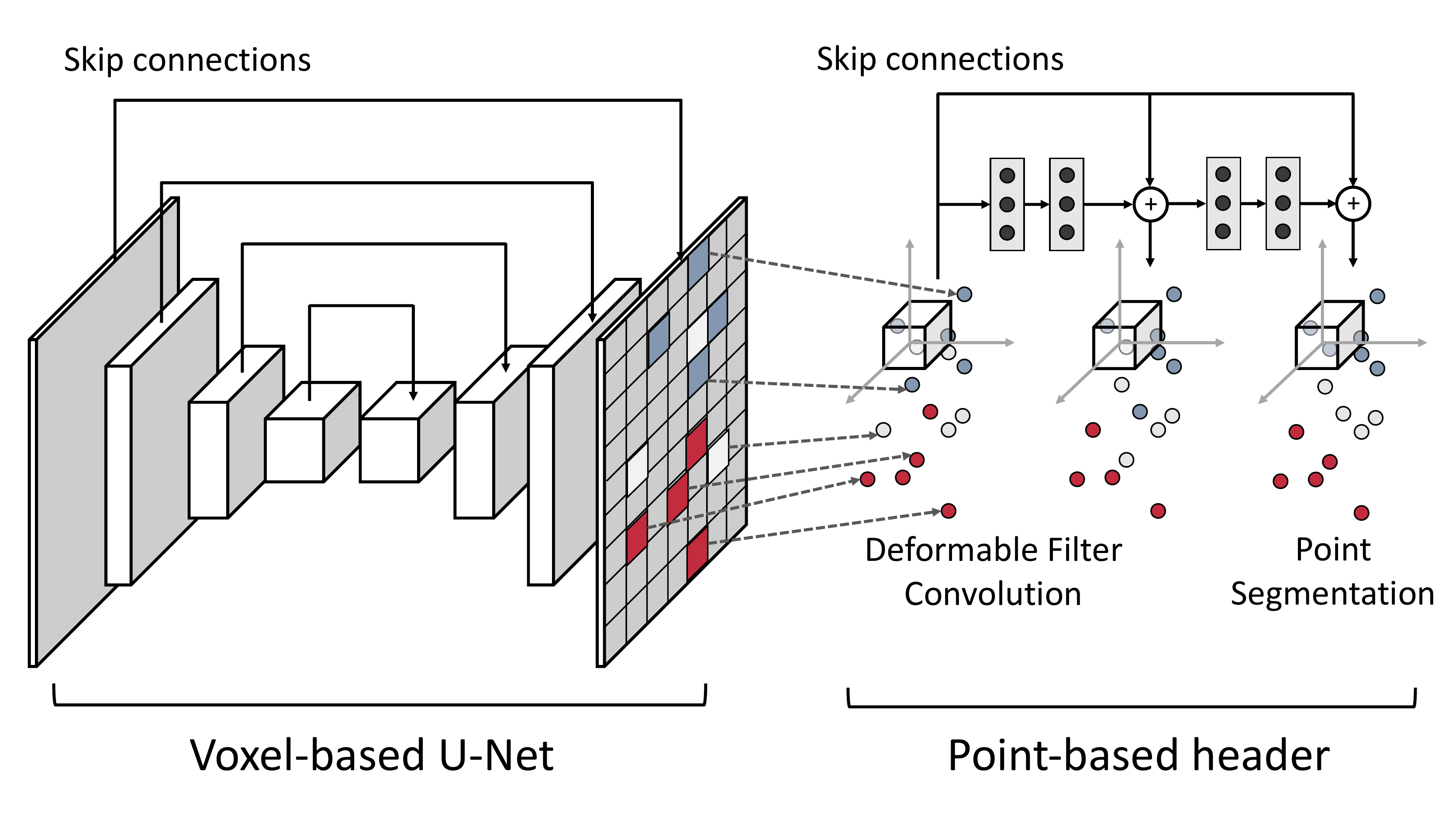}
\caption{Our point cloud segmentation network contains a voxelized backbone U-Net and a point-based
header using our proposed deformable filter convolution layers.}
\label{fig:unet}
\end{figure}

\paragraph{Point cloud convolution using voxel features:} We adopt a UNet architecture \cite{unet}
to extract voxelized features and then add our point convolution layer on top. One baseline approach
is to use a voxelized network only, which suffers from the precision of the output, since points
belonging to the same voxel will have the same output. To predict on the point level, one needs to
fuse the voxelized feature onto point clouds. We consider the continuous fusing operator (ContFuse)
proposed in \cite{contfuse} as a strong baseline. This operator is similar to what has been done in
PointNet++ \cite{pointnet2}, where each point first queries a local neighborhood, then passes the
neighboring point features plus coordinate offsets through an MLP, and finally averages the
neighborhood features together. Note that our approach is much more memory efficient than the
ContFuse operation, since we do not need to tile the neighboring point features into a larger size
tensor. PCC \cite{pccn} also adopts the same setting where they found the best performance when their
continuous convolution layers are composed on top of the voxelized features. Figure~\ref{fig:unet}
illustrates the overall network architecture, where the point-based convolution header takes inputs
from a voxelized feature extractor.

\begin{figure}[t]
\centering
\includegraphics[width=\columnwidth]{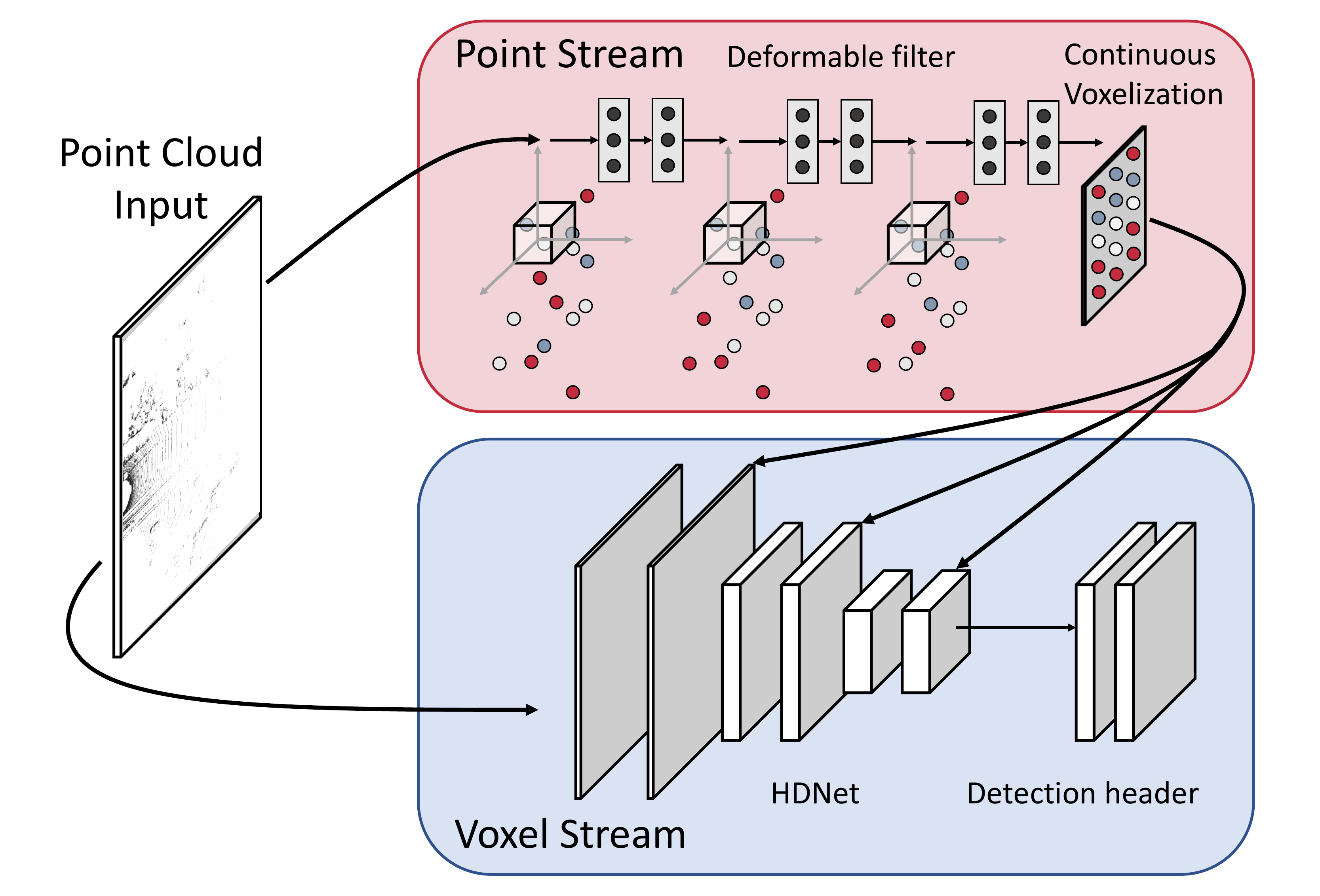}
\caption{Our point cloud detection network fuses point features into various stages of the voxelized
backbone network.}
\label{fig:hdnetfuse}
\end{figure}
\vspace{-0.1in}

\paragraph{Implementation details:}
The voxel feature extractor uses a standard U-Net \cite{unet} with [32, 64, 128, 256] channels in
the encoder and decorder. Each encoder block contains convolution, batch norm, ReLU, and max pooling
layers. The decoder has a symmetric structure, with max-pooling replaced by bilinear interpolation
to upsample the feature map. Skip connections are added between each pair of corresponding layers
connecting the encoder and decoder. Similar to~\cite{pccn}, the ContFuse baseline has 7 blocks in
the point-based header, each containing an MLP of size [11, 8, 8, 8], where the inputs are
concatenated with the offset coordinates of the neighborhood (3-dimensional) and the output classes
(7 plus none of the above). There are skip connections combining the outputs of these blocks. Our
deformable version has 2 blocks, each with [8, 16, 32] channels, and skip connections are also applied
on the output channels. We used 3$\times$3$\times$3 convolution filters with filter grid unit length
0.2m and neighborbood size 16. We used the Adam optimizer with initial learning rate 1e-4, weight
decay 5e-4, batch size 16 and 0.1$\times$ learning rate decay at 50k and 100k iterations. The
baseline U-Net was trained for 450k iterations, and ContFuse and our deformable filter version were
trained for 115k iterations.

\begin{figure*}[t]
\vspace{0.2in}
\centering
\begin{minipage}{\textwidth}
\centering
\footnotesize
\begin{tabular}{c c c c}
\includegraphics[width=0.3\textwidth,trim={20cm 10cm 20cm 5cm},clip]{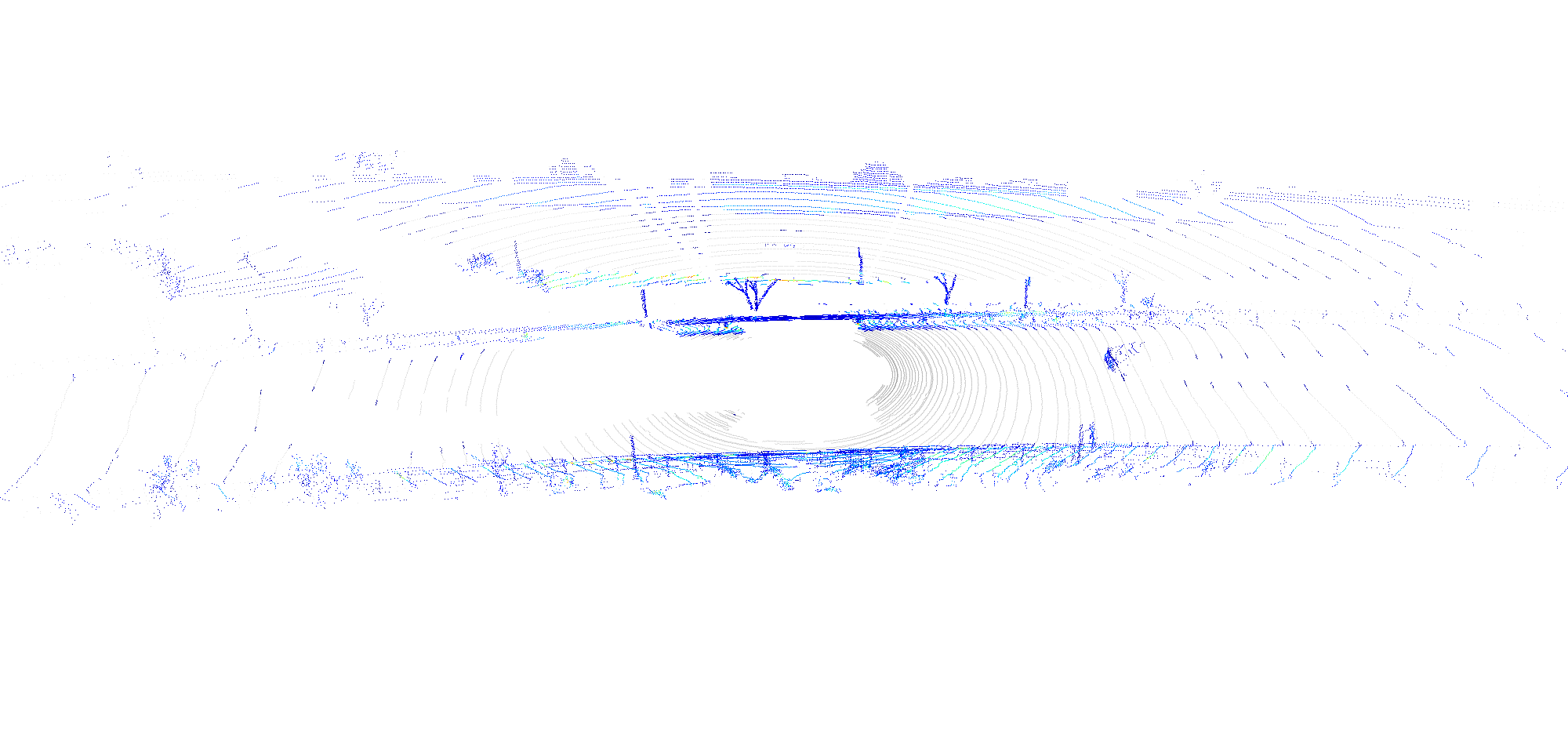} &
\includegraphics[width=0.3\textwidth,trim={20cm 10cm 20cm 5cm},clip]{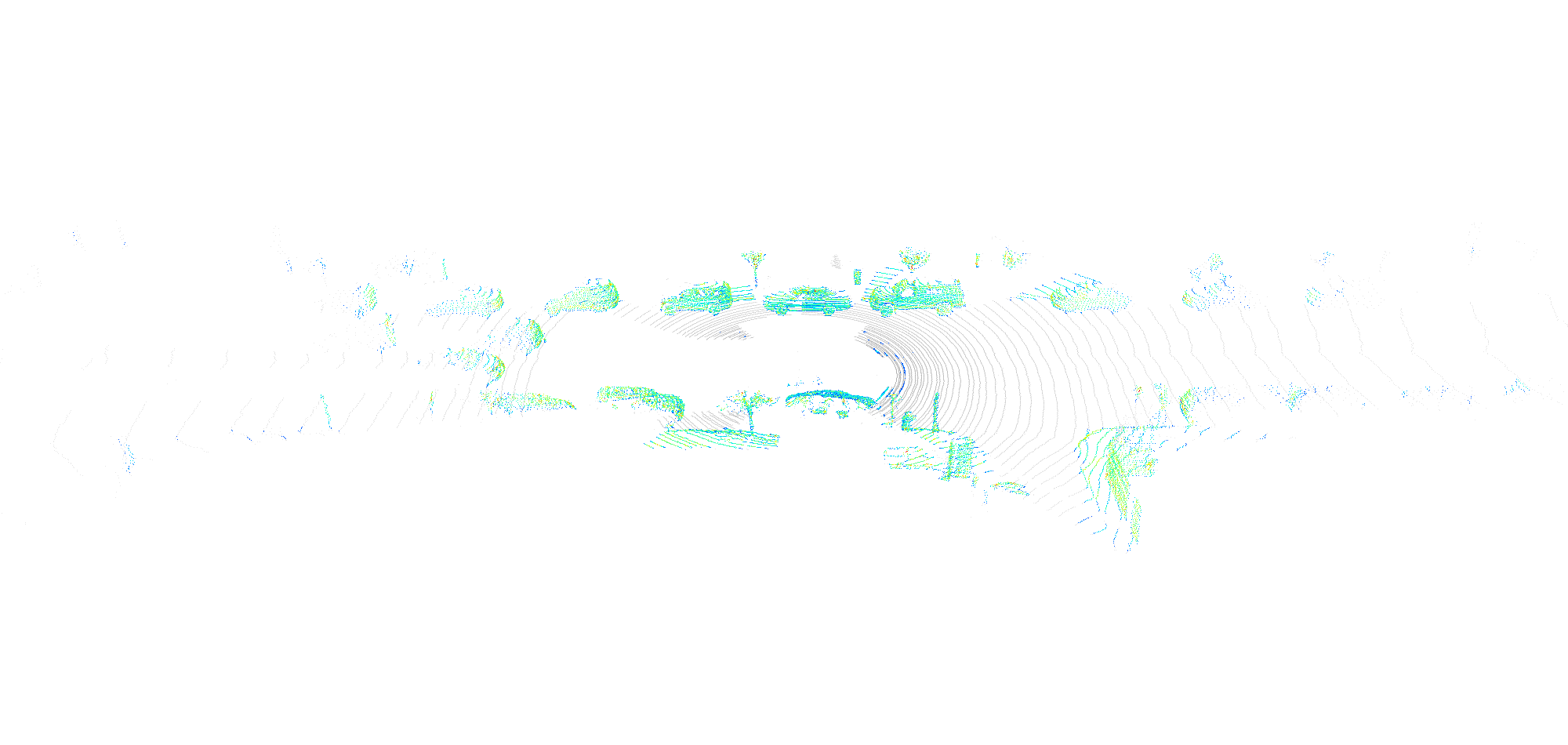}&
\includegraphics[width=0.3\textwidth,trim={20cm 10cm 20cm 5cm},clip]{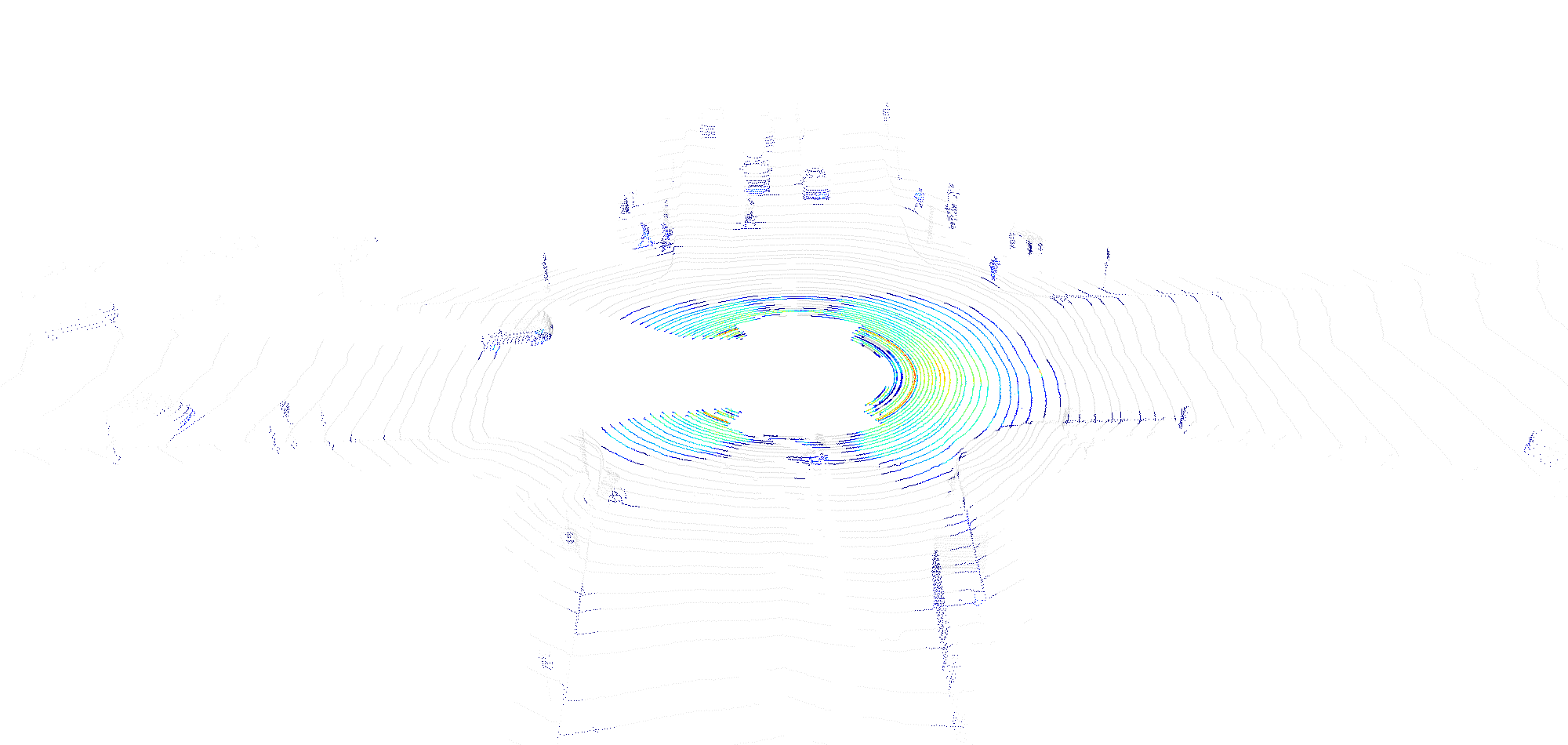}&
\end{tabular}
\end{minipage}
\vspace{0.05in}
\caption{Visualization of different learned filter channels using guided backprop}
\label{fig:guidedbp}
\end{figure*}

\begin{figure*}[h!]
\vspace{-0.2in}
\centering
\begin{minipage}{\textwidth}
\centering
\footnotesize
\begin{tabular}{c c c}
\fbox{\includegraphics[width=0.3\textwidth,trim={15cm 10cm 25cm 5cm},clip]{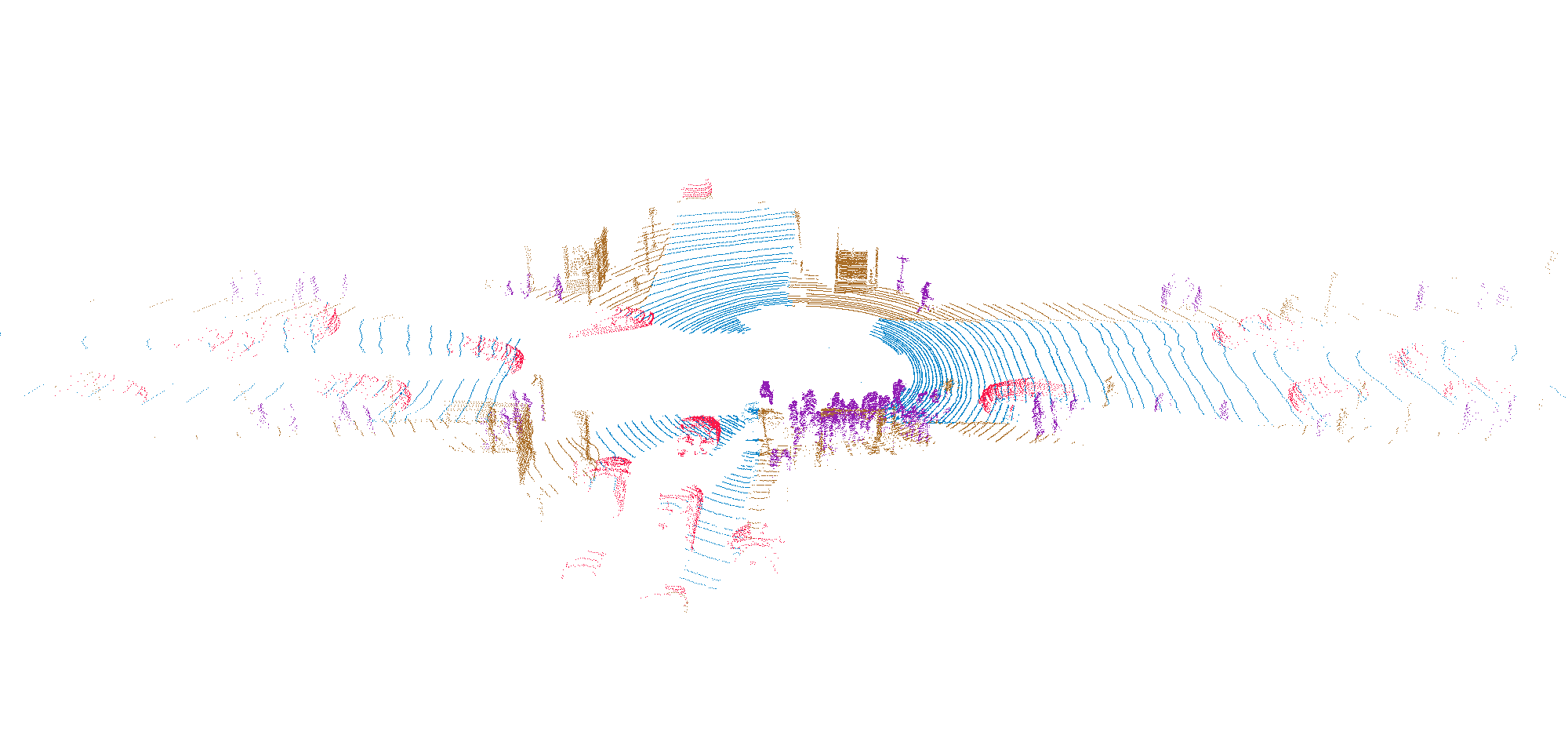}} &
\fbox{\includegraphics[width=0.3\textwidth,trim={15cm 10cm 25cm 5cm},clip]{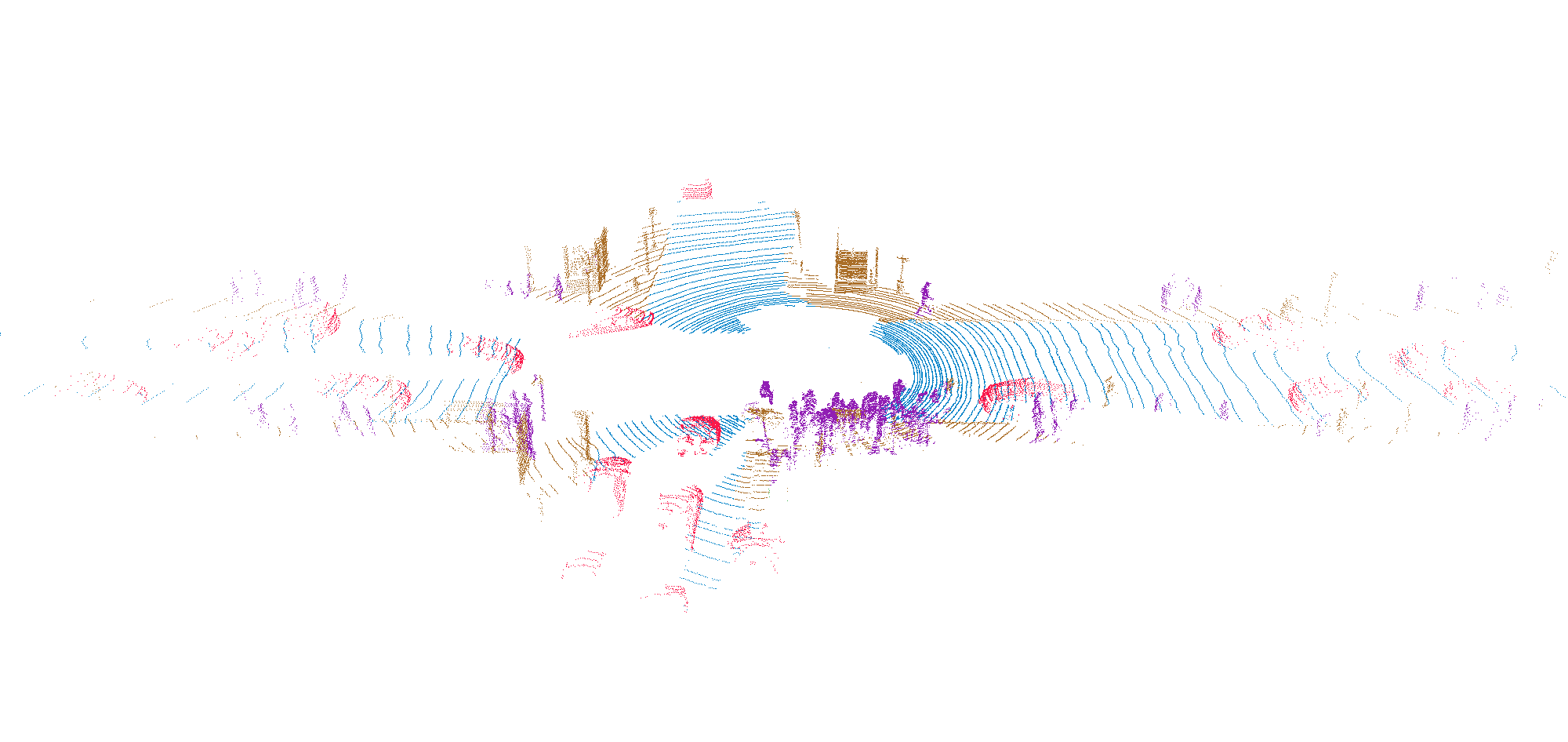}} &
\fbox{\includegraphics[width=0.3\textwidth,trim={15cm 10cm 25cm 5cm},clip]{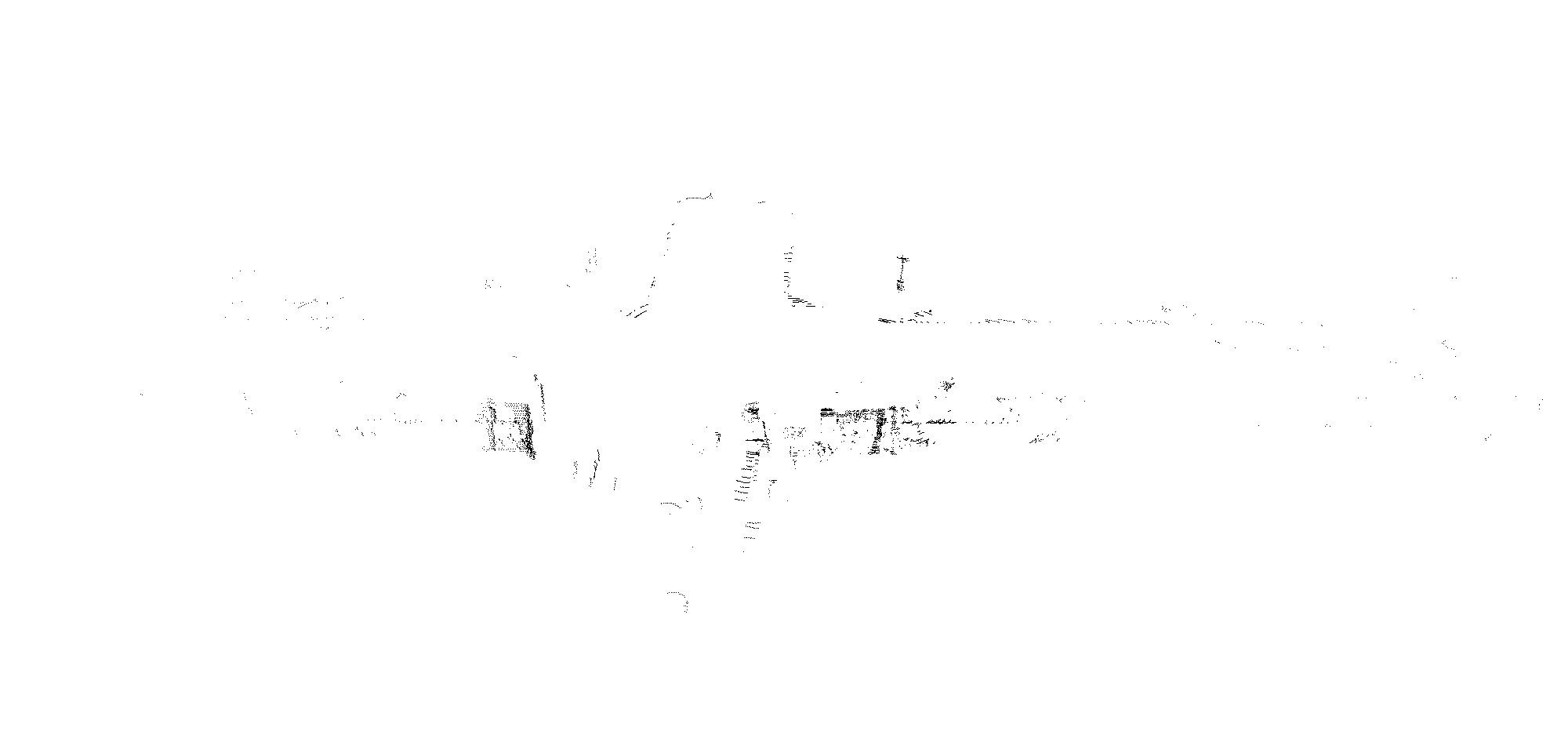}}\\
GT & U-Net & U-Net Error \\
&
\fbox{\includegraphics[width=0.3\textwidth,trim={15cm 10cm 25cm 5cm},clip]{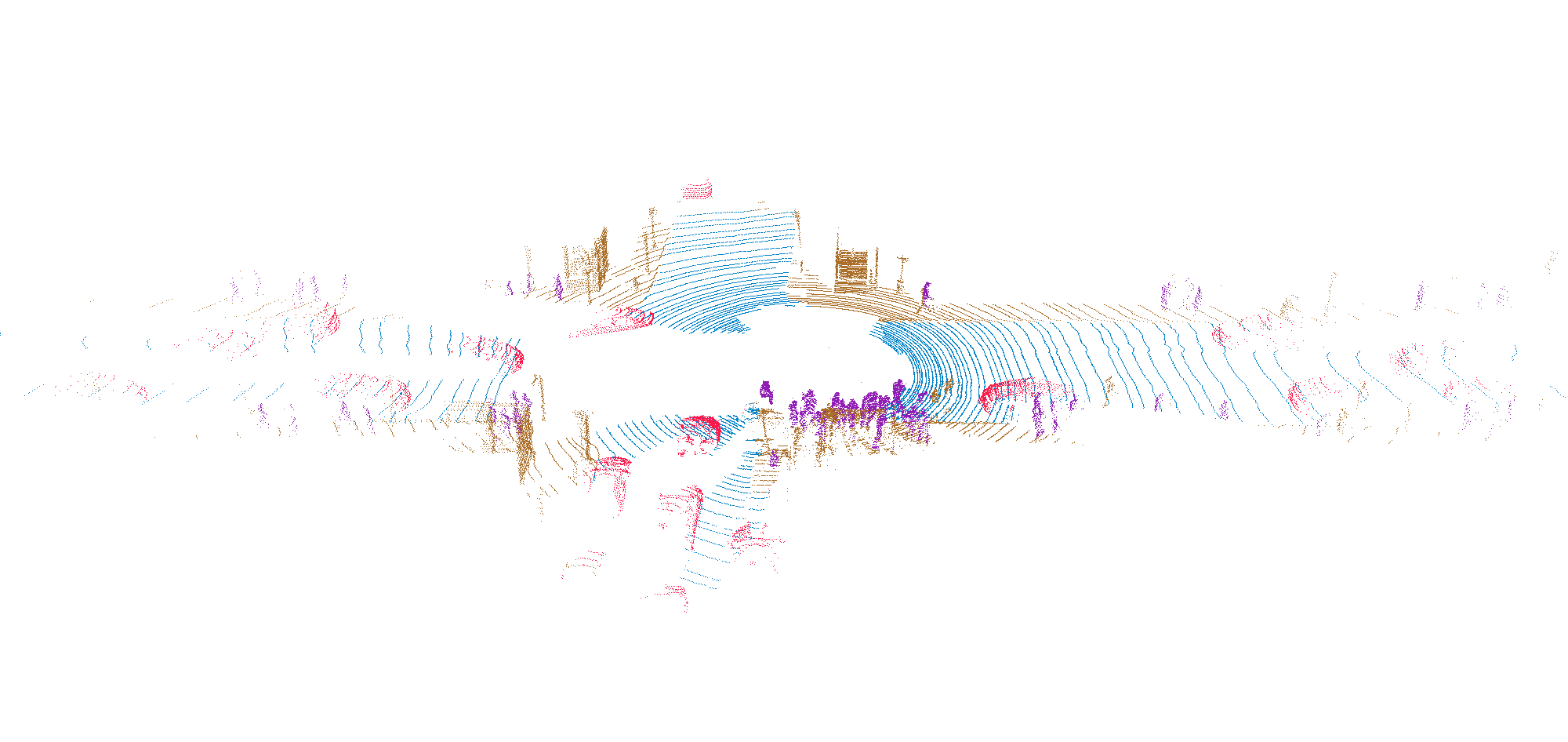}}
&
\fbox{\includegraphics[width=0.3\textwidth,trim={15cm 10cm 25cm 5cm},clip]{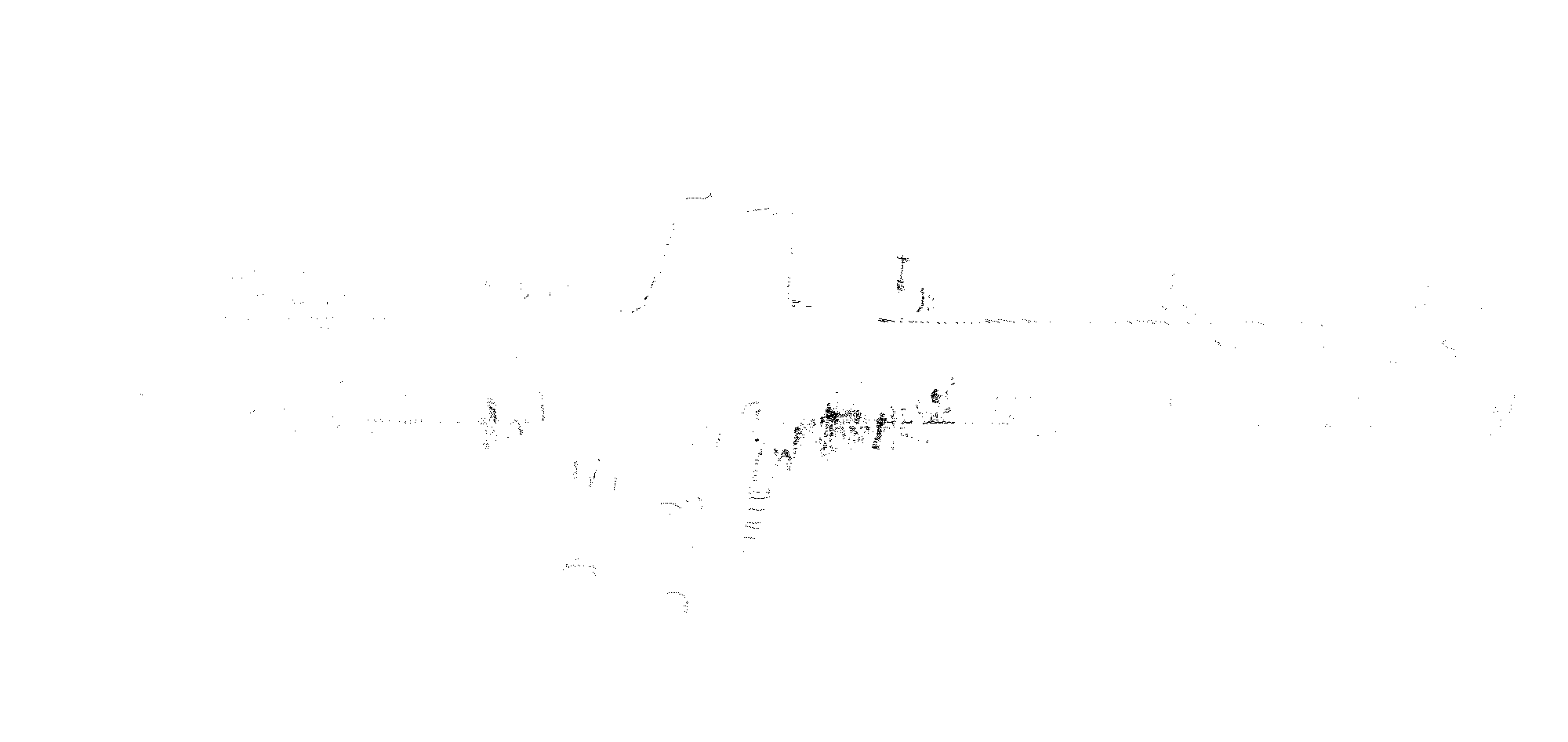}}
\\
& U-Net +DeformFilter & U-Net +DeformFilter Error \\
\\
\\
\fbox{\includegraphics[width=0.3\textwidth,trim={15cm 10cm 25cm 5cm},clip]{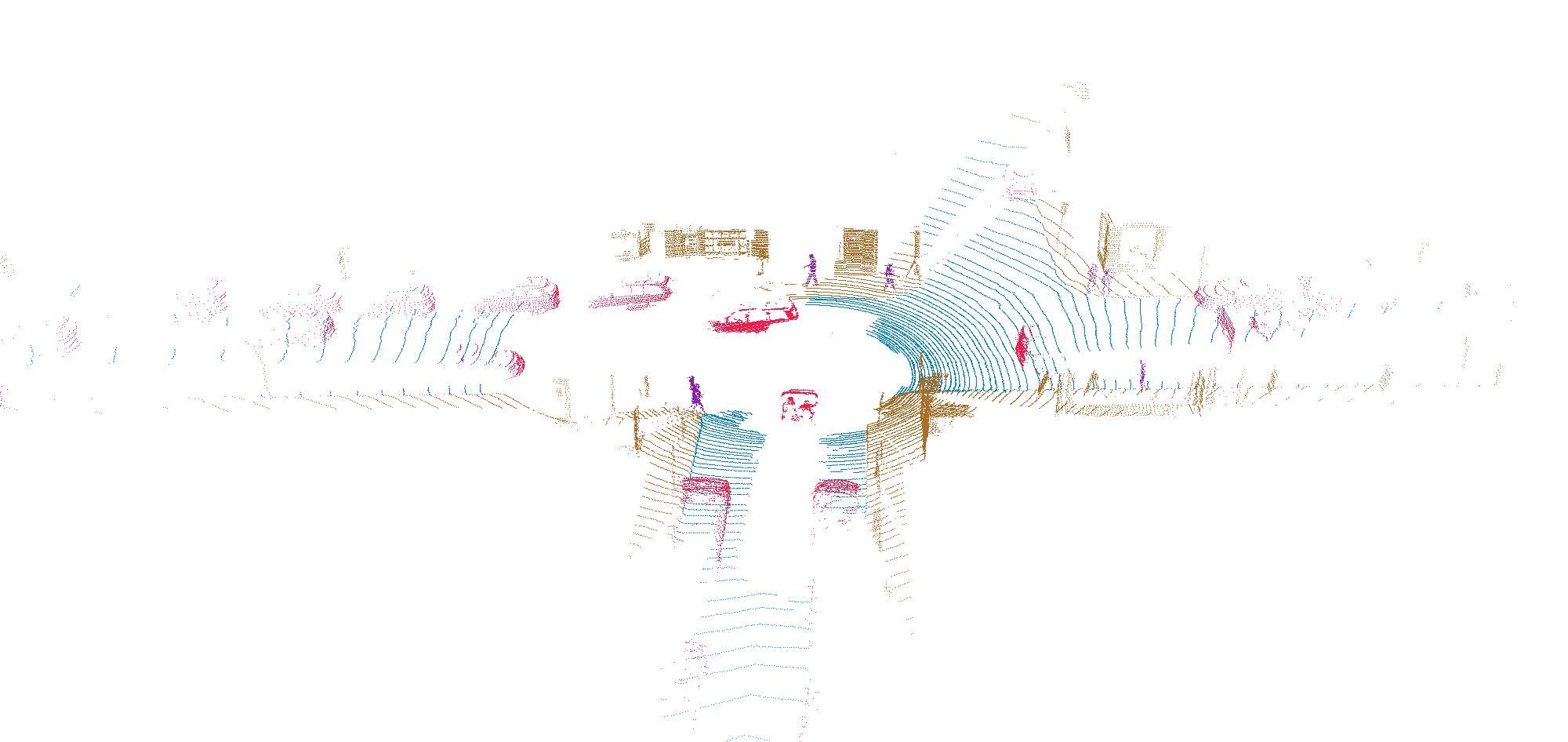}} &
\fbox{\includegraphics[width=0.3\textwidth,trim={15cm 10cm 25cm 5cm},clip]{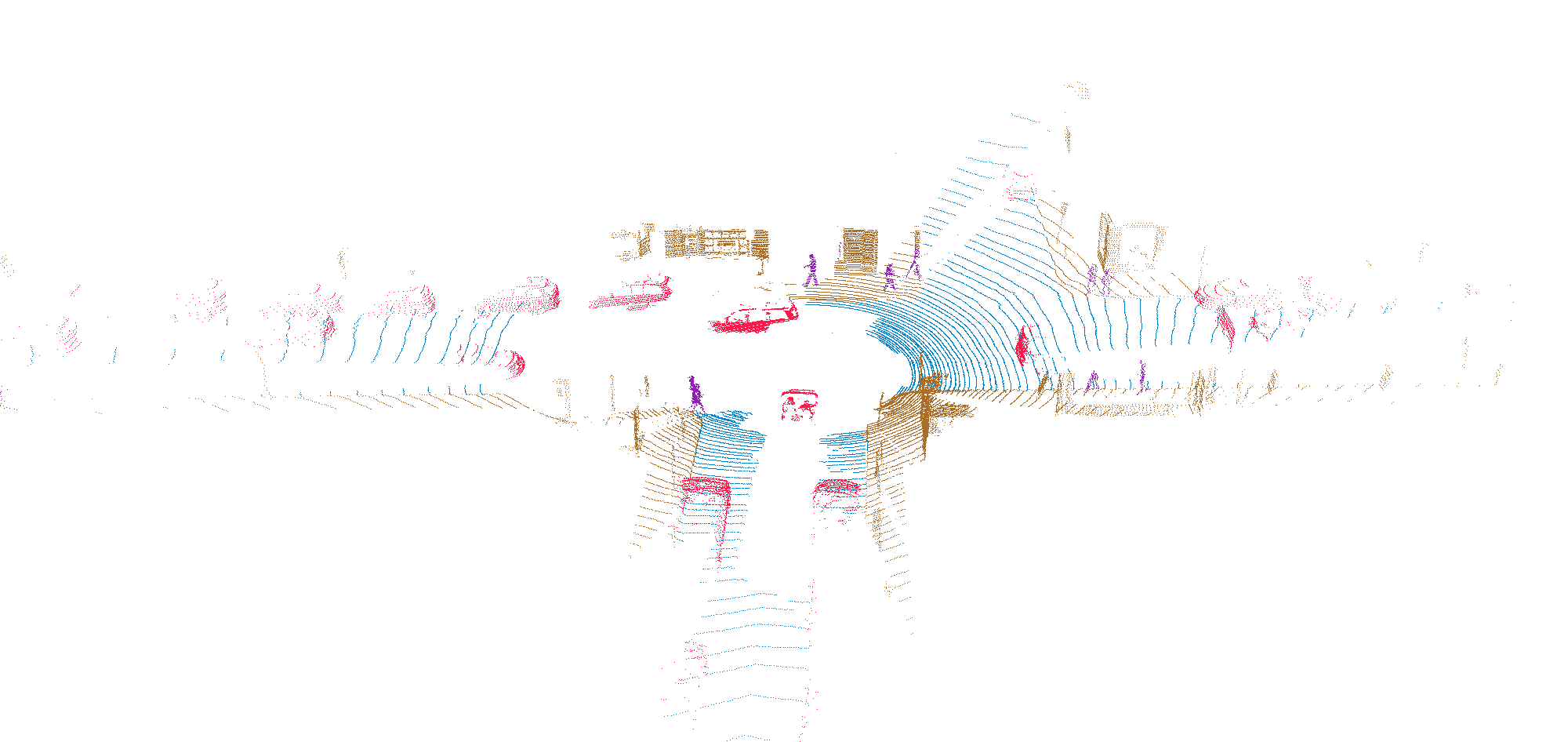}} &
\fbox{\includegraphics[width=0.3\textwidth,trim={15cm 10cm 25cm 5cm},clip]{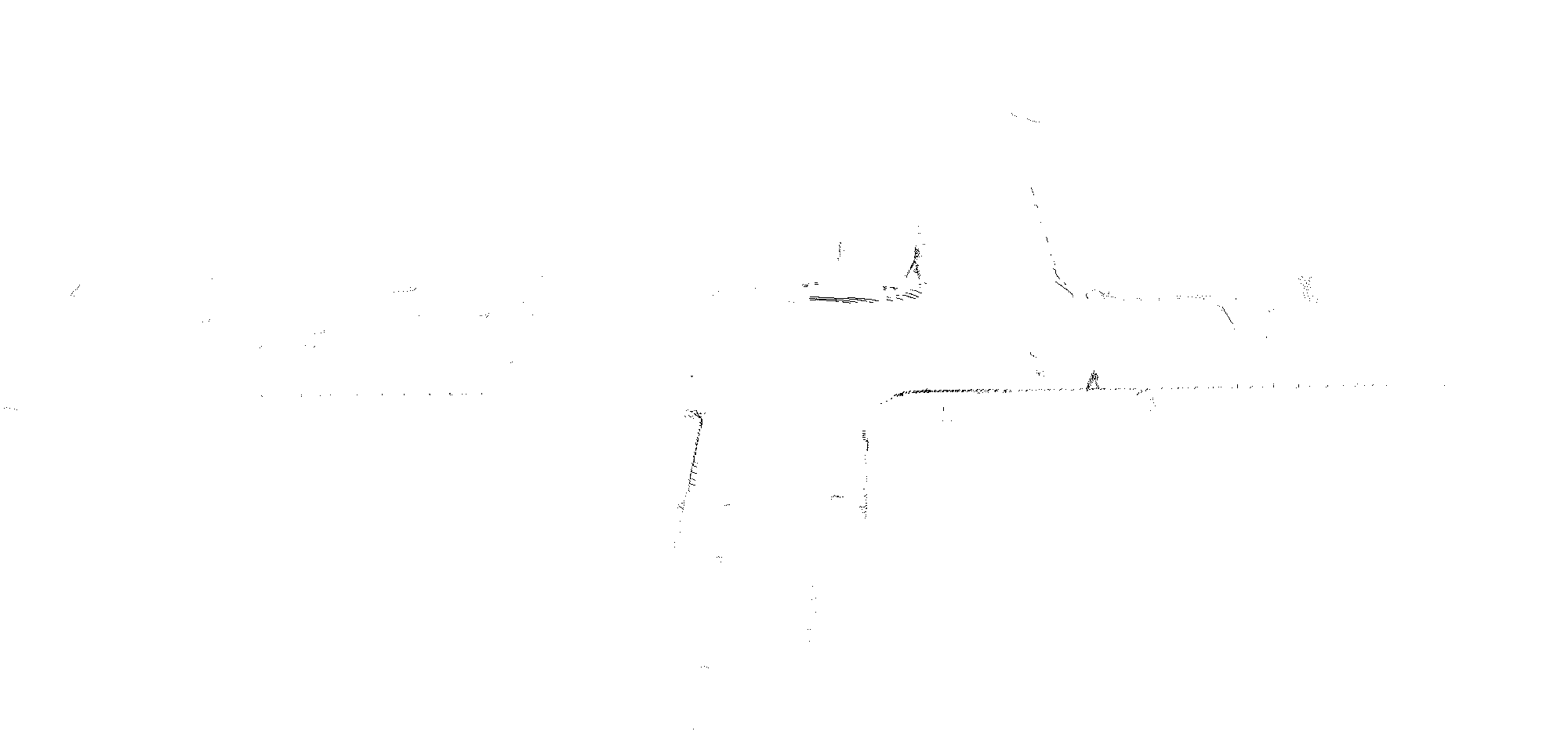}} \\
GT & U-Net & U-Net Error \\
&
\fbox{\includegraphics[width=0.3\textwidth,trim={15cm 10cm 25cm 5cm},clip]{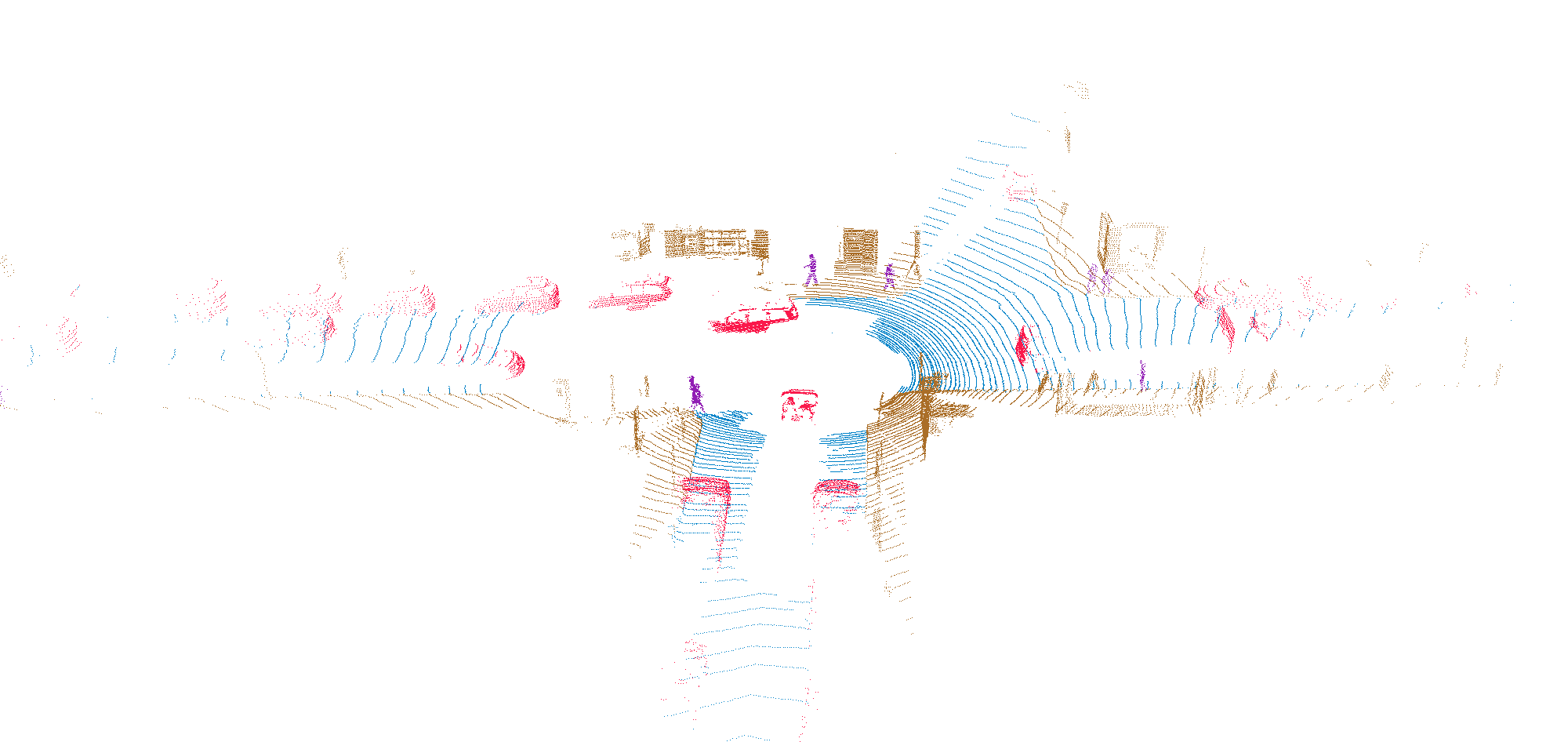}} &
\fbox{\includegraphics[width=0.3\textwidth,trim={15cm 10cm 25cm 5cm},clip]{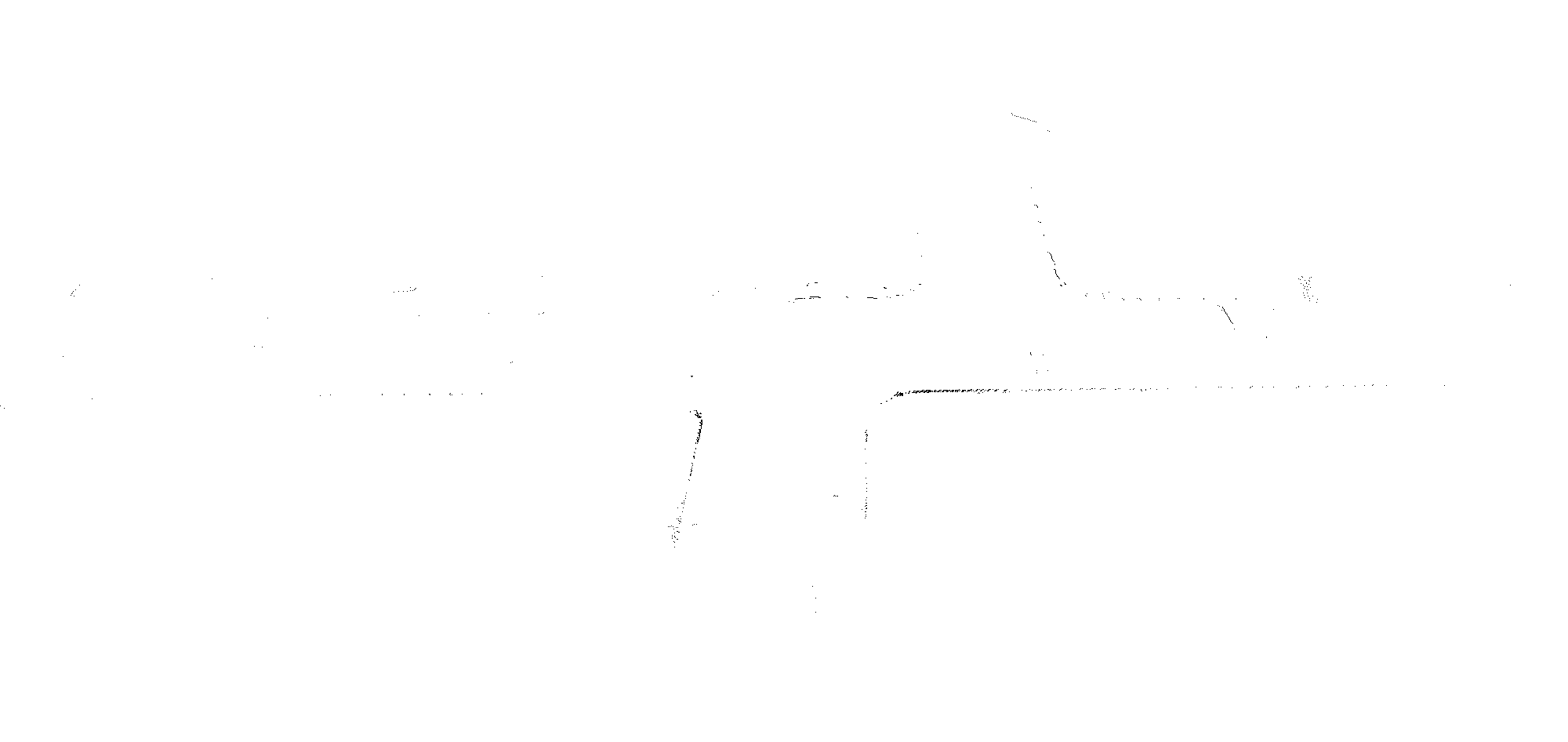}} \\
& U-Net +DeformFilter & U-Net +DeformFilter Error \\
\\
\\
\fbox{\includegraphics[width=0.3\textwidth,trim={15cm 10cm 25cm 5cm},clip]{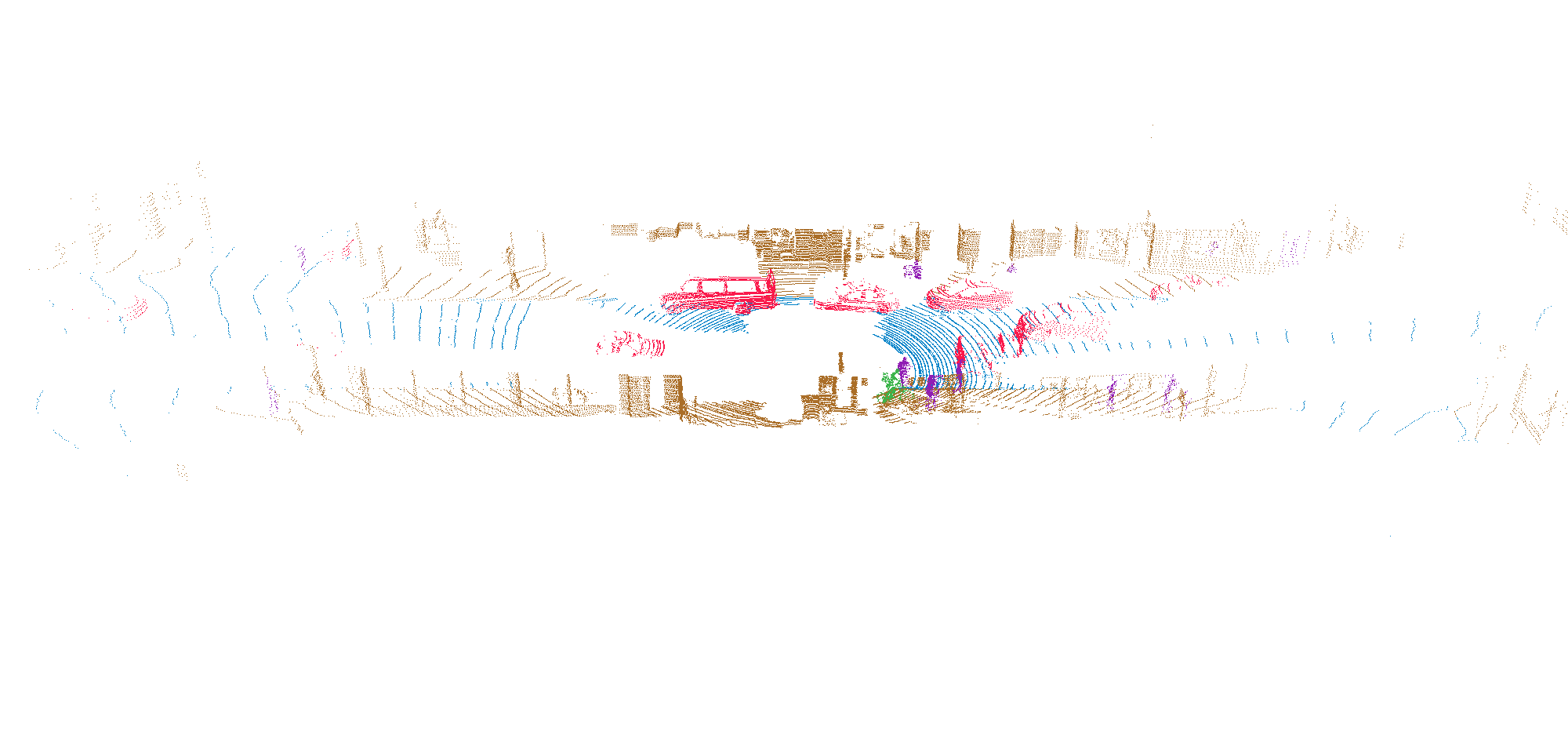}} &
\fbox{\includegraphics[width=0.3\textwidth,trim={15cm 10cm 25cm 5cm},clip]{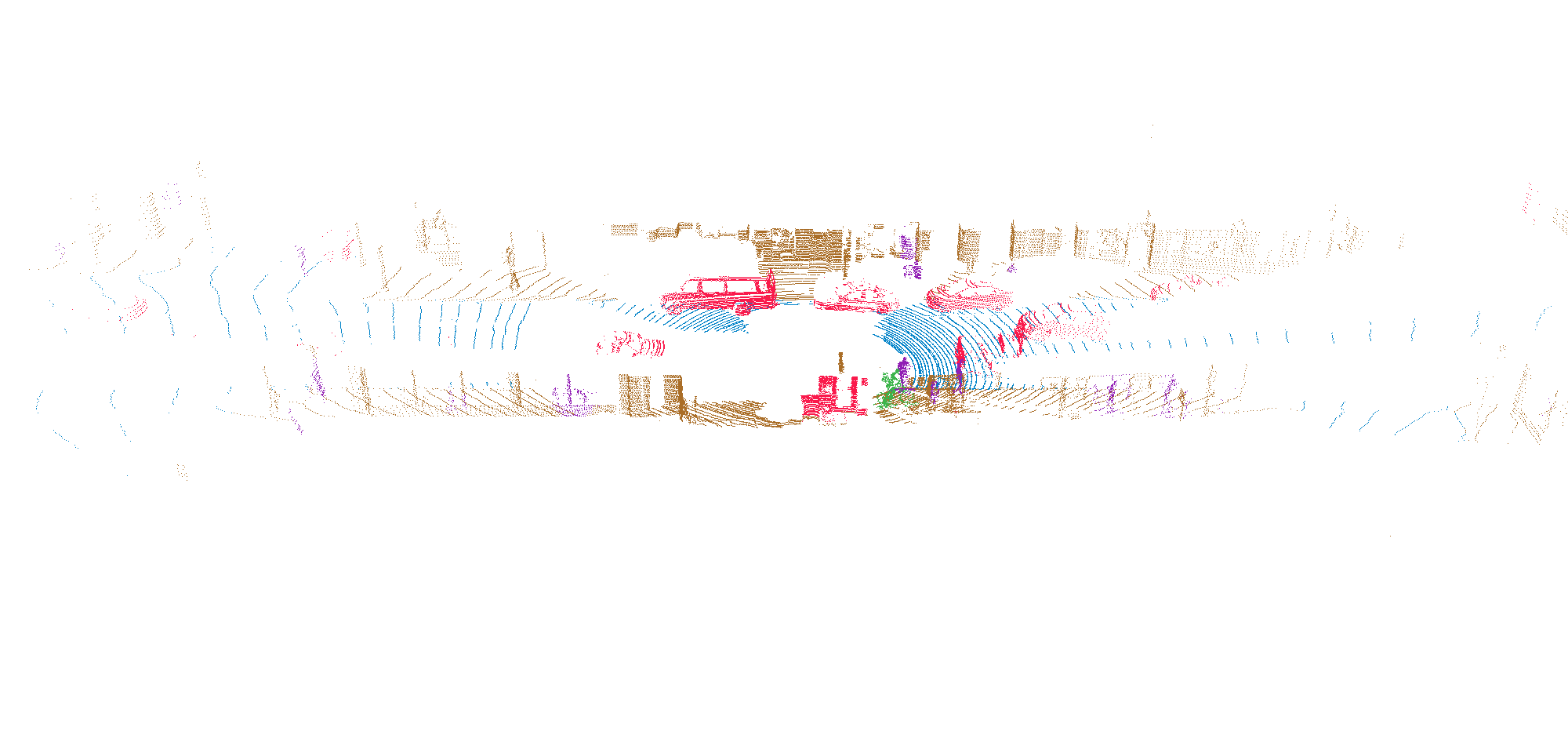}} &
\fbox{\includegraphics[width=0.3\textwidth,trim={15cm 10cm 25cm 5cm},clip]{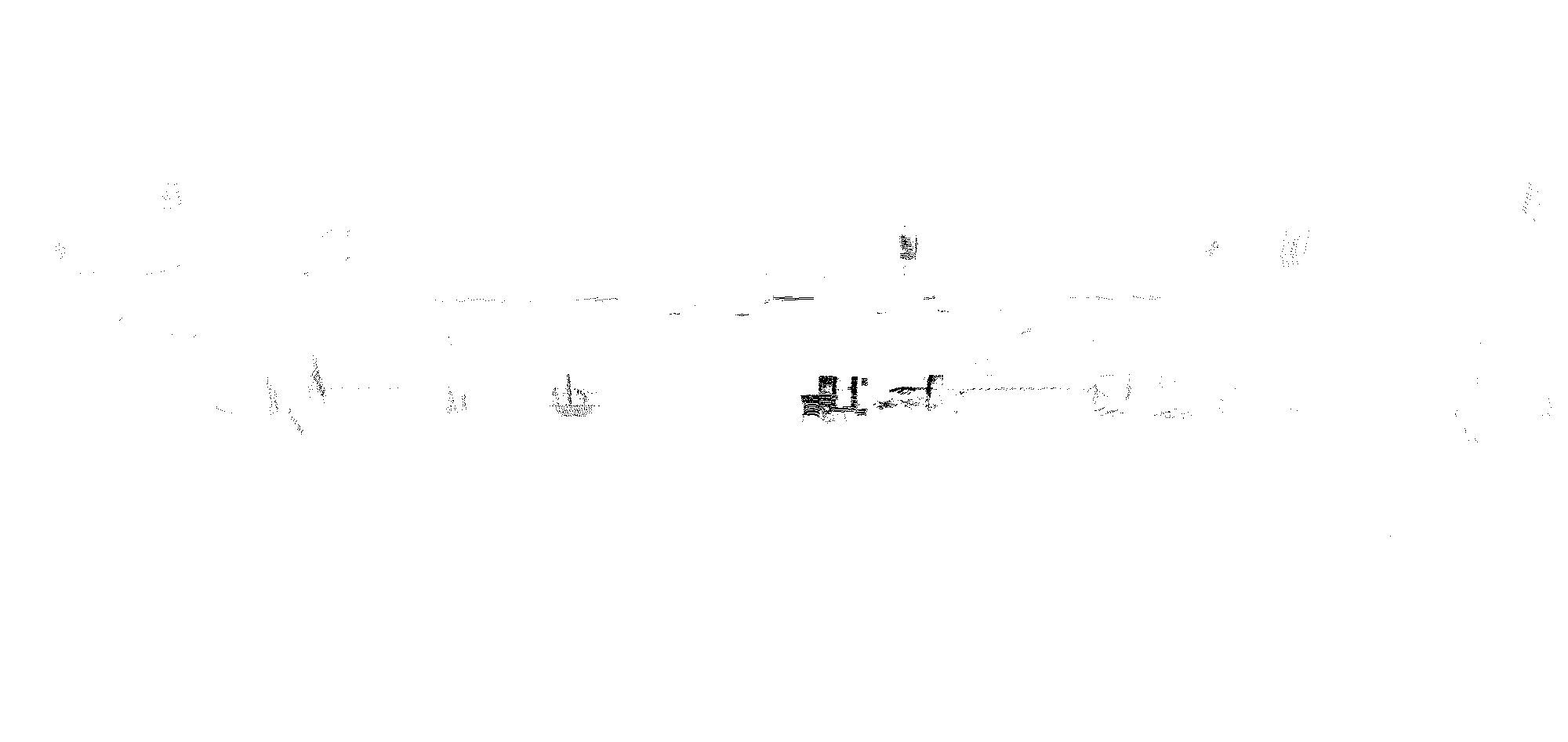}}\\
GT & U-Net &  U-Net Error \\
&
\fbox{\includegraphics[width=0.3\textwidth,trim={15cm 10cm 25cm 5cm},clip]{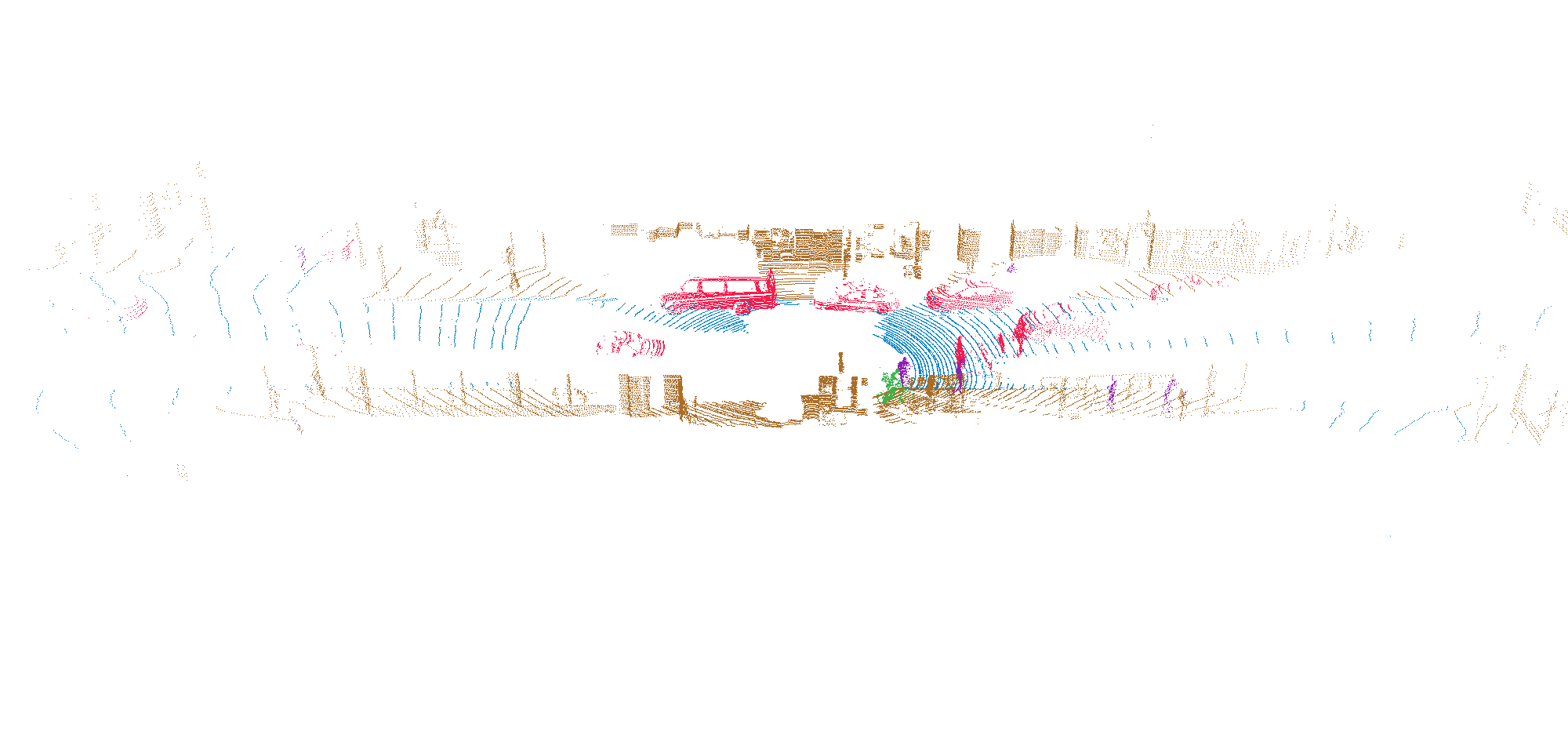}} &
\fbox{\includegraphics[width=0.3\textwidth,trim={15cm 10cm 25cm 5cm},clip]{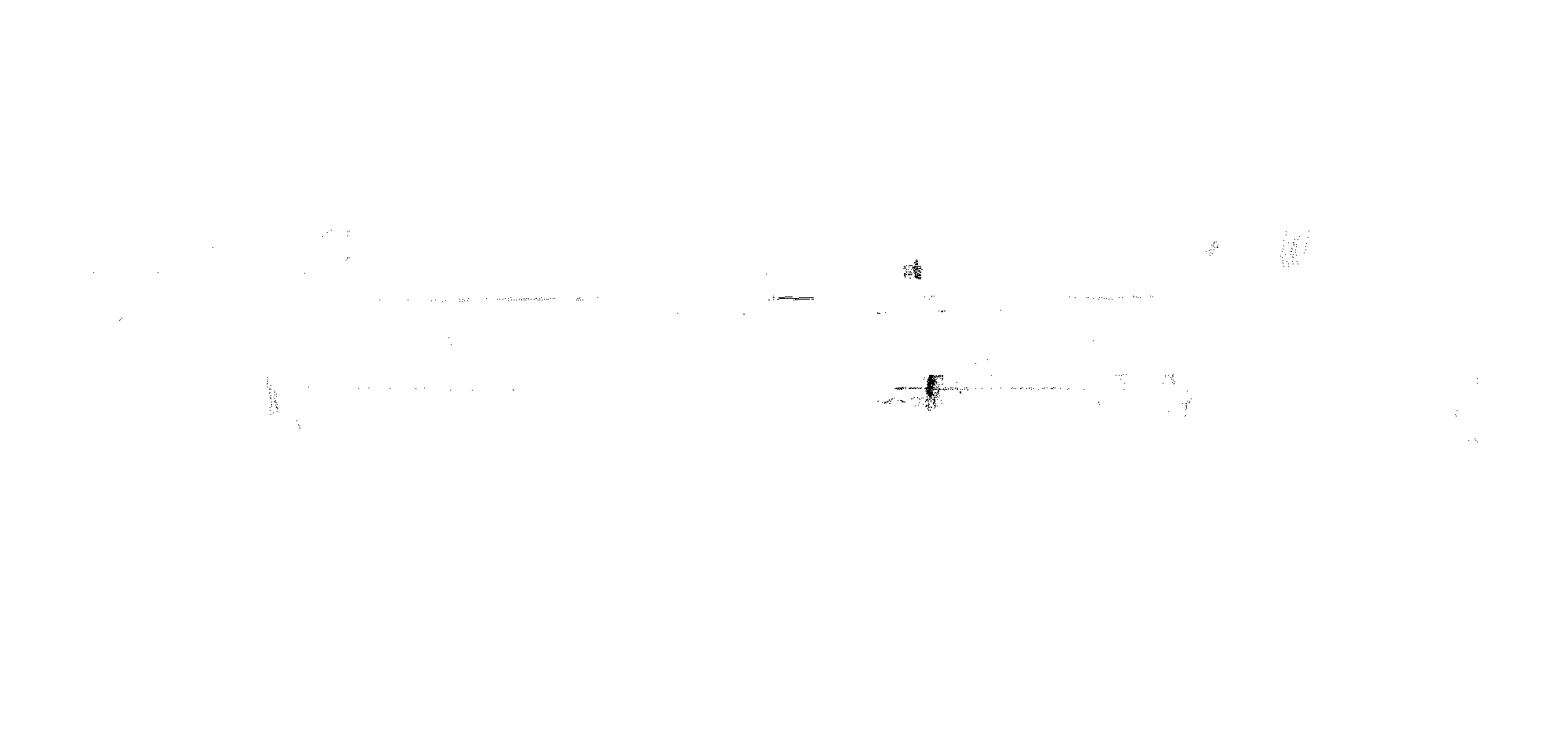}} \\
& U-Net +DeformFilter & U-Net +DeformFilter Error \\
\end{tabular}
\end{minipage}
\vspace{0.05in}
\caption{Visualization of results on TOR4D LiDAR semantic segmentation}
\label{fig:semantic}
\end{figure*}

\paragraph{Results and discussion:}
Results on TOR4D test set are shown in Table~\ref{tab:semseg} and qualitative visualization of the
output results shown in Figure~\ref{fig:semantic}. Our joint network using deformable filter layers
signifcantly surpasses the baseline, achieving state-of-the-art performance. Notably, our deformable
filter model outperforms the ContFuse baseline, which uses an MLP to aggregate point cloud
neighborhood information. To understand filter activations, we also plot the most activated region
of each filter channel using guided backprop~\cite{guidedbp}, shown in Figure~\ref{fig:guidedbp}. It
is clear that the regions that activate individual neurons the most are roughly corresponding to the
semantic classes.

\begin{figure*}[t]
\vspace{-0.18in}
\centering
\begin{minipage}{\textwidth}
\centering
\begin{tabular}{c c c}
\multirow{2}{*}{PCC}             & 
\includegraphics[width=0.45\textwidth,trim={0 0.8cm 0 0.8cm},clip]{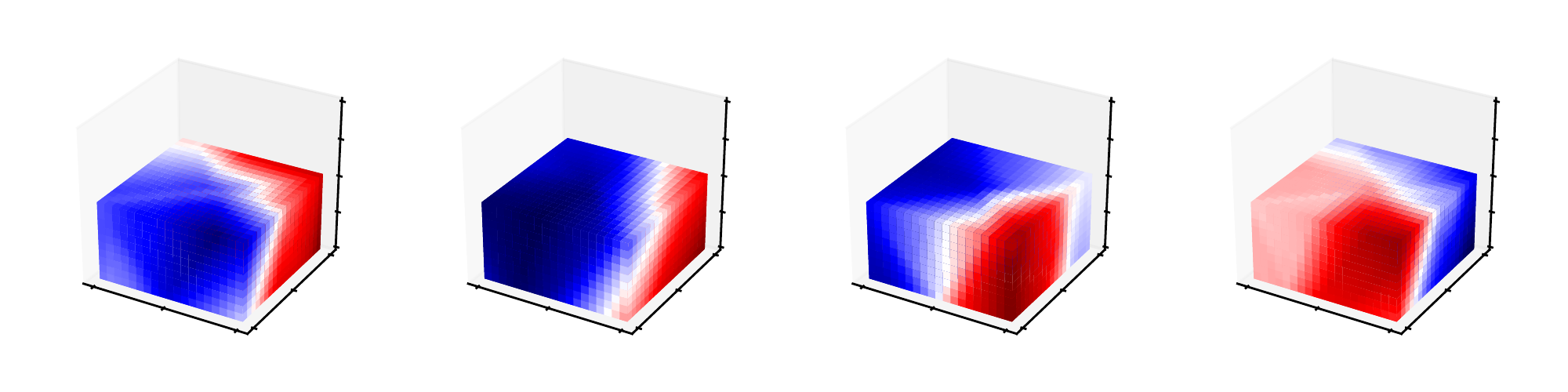} & 
\includegraphics[width=0.45\textwidth,trim={0 0.8cm 0 0.8cm},clip]{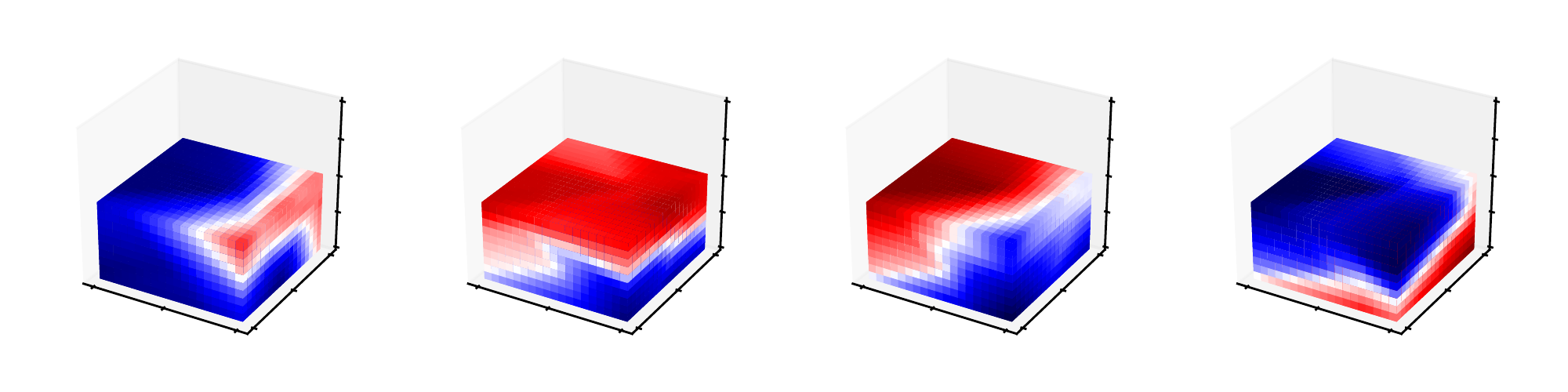} \\
                                 & 
\includegraphics[width=0.45\textwidth,trim={0 0.8cm 0 0.8cm},clip]{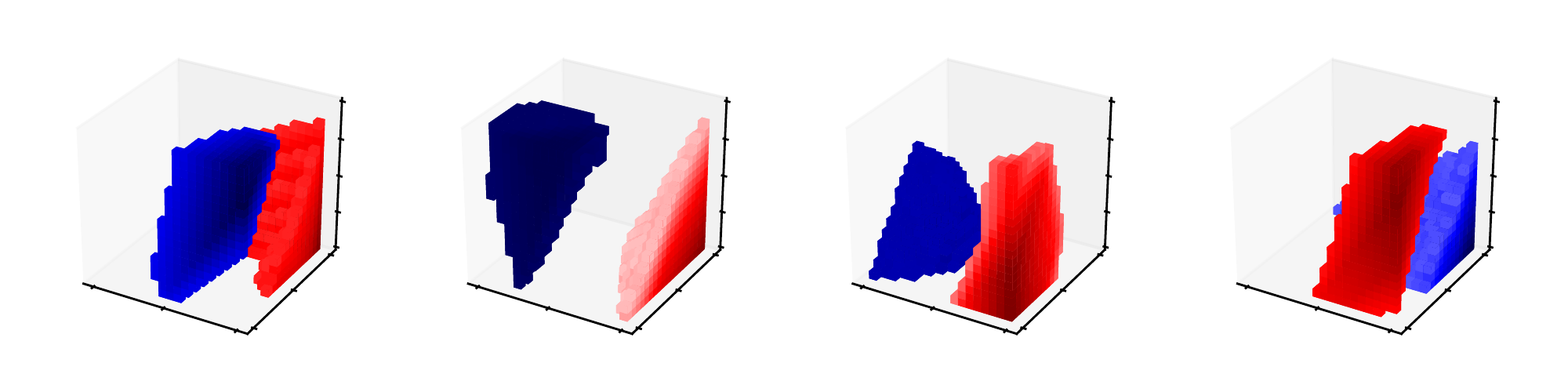} & 
\includegraphics[width=0.45\textwidth,trim={0 0.8cm 0 0.8cm},clip]{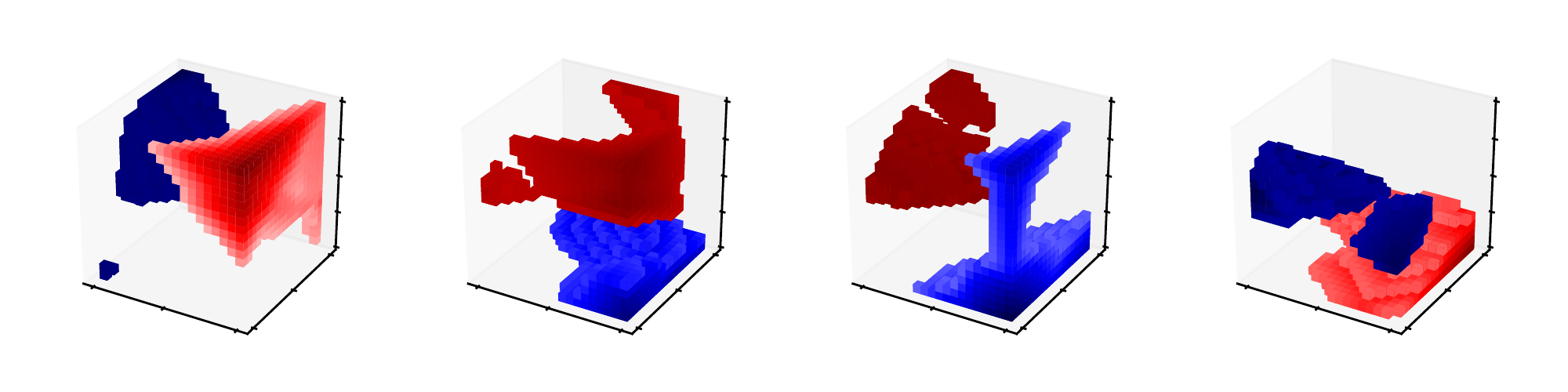} \\
                                 & {\footnotesize Layer 1}                                                        
                                 & {\footnotesize Layer 2}                                           \\
                                 \\
\multirow{2}{*}{Ours}            & 
\includegraphics[width=0.45\textwidth,trim={0 0.8cm 0 0.8cm},clip]{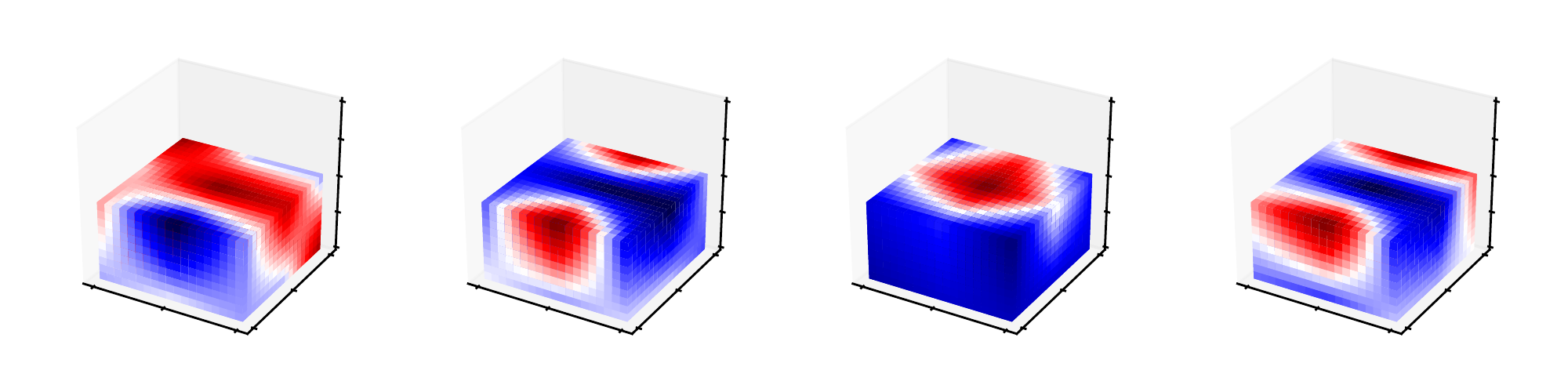}     & 
\includegraphics[width=0.45\textwidth,trim={0 0.8cm 0 0.8cm},clip]{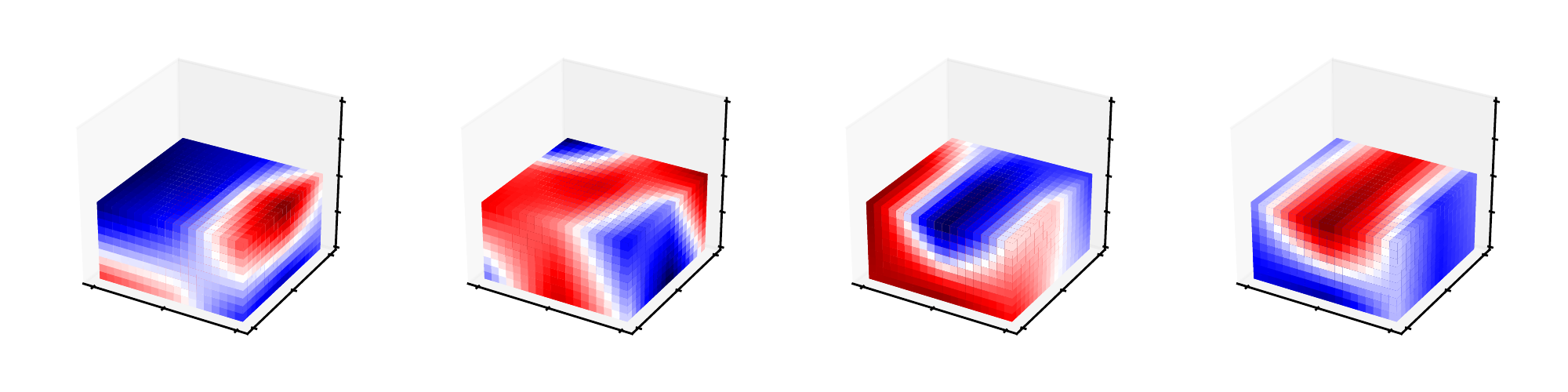}     \\
                                 & 
\includegraphics[width=0.45\textwidth,trim={0 0.8cm 0 0.8cm},clip]{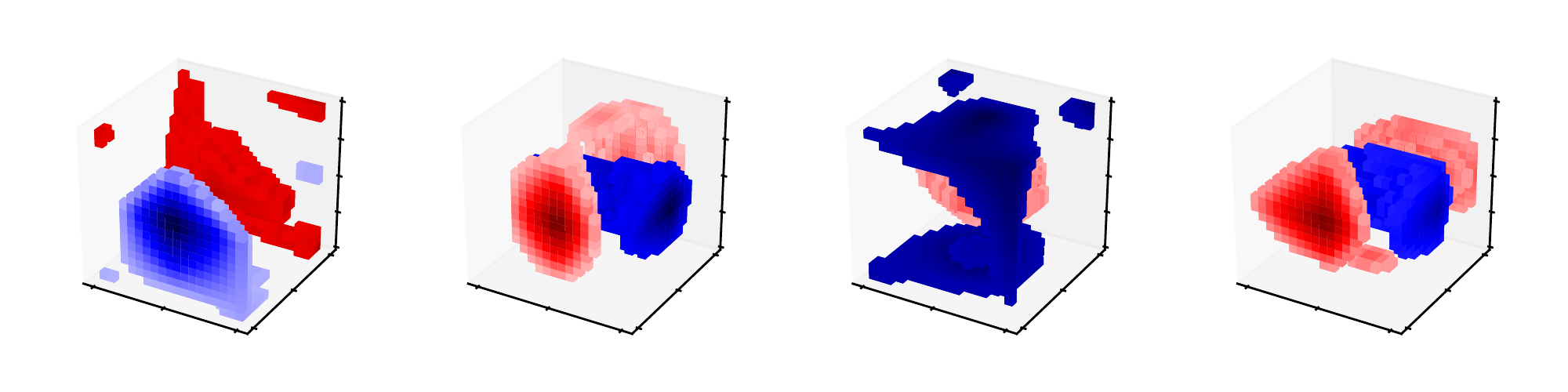}     & 
\includegraphics[width=0.45\textwidth,trim={0 0.8cm 0 0.8cm},clip]{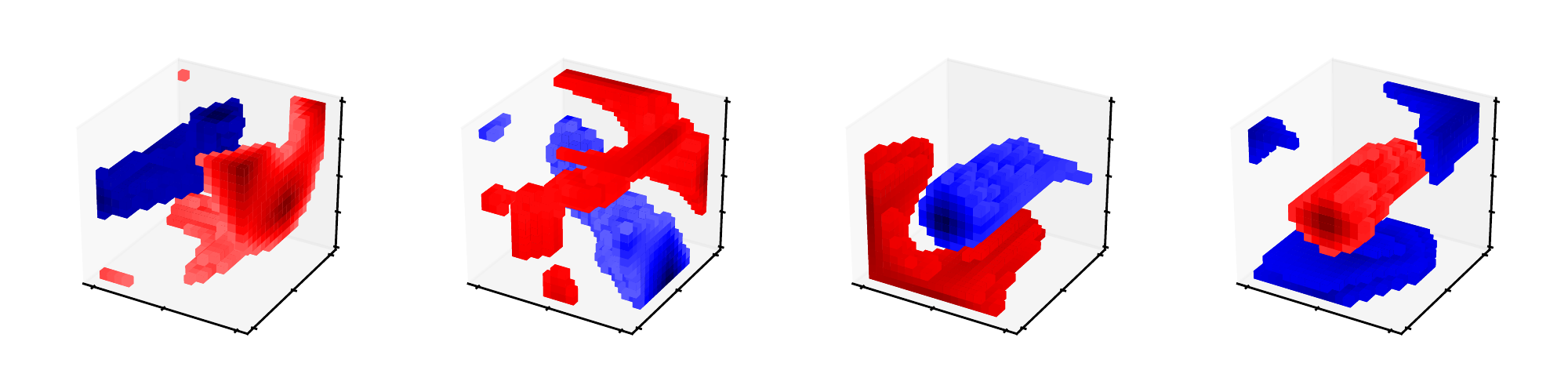}     \\
                                 & {\footnotesize Layer 1}                                                        
                                 & {\footnotesize Layer 2}                                           \\
\end{tabular}
\end{minipage}
\vspace{0.05in}
\caption{Visualization of our learned 3D filters ($7 \times 7 \times 7$) trained on KITTI cyclist
detection, compared to filters output by an MLP (PCC) \cite{pccn}. The upper row is the full filter
dissected by half across $z$-axis; the lower row is the top and bottom 10\% quantile of the filter
weights. \textcolor{red}{Red} denotes positive filter weights and \textcolor{blue}{blue} negative.
}
\label{fig:filterviz}
\end{figure*}

\subsection{KITTI BEV object detection}
We evaluate our deformable filter convolution on KITTI BEV object detection benchmark \cite{kitti},
consisting of 7,481 training and 7,518 testing frames of LiDAR point clouds. Half of the training
data is split for validation. Detectors on ``car'', ``pedestrian'' and ``cyclist'' categories are
trained and evaluated individually. 

We adopt the HDNet architecture proposed in \cite{hdnet}, one of the top-performing object detectors
on this benchmark. HDNet is a voxelized network that performs regular 2D convolution on BEV with the
$z$-axis to be the channel dimension. It also predicts the map information with a pretrained module.
To preserve local geometric details and to provide extra information to the voxelized backbone
network, we add a deformable filter input branch that processes raw LiDAR points and the output of
the branch is fused with the backbone network. Shown in Figure~\ref{fig:hdnetfuse}, we add 3 layers
of deformable filter convolution to process the point cloud, with channel dimension [4, 16, 32]
respectively. The last deformable filter layer samples the output points at the voxel centers so
that we can concatenate the feature with the voxel input branch. We compare our proposed operator
with parametric continuous convolution (PCC) \cite{pccn}, which uses an MLP to predict the
convolution filter weights. The MLP takes 3-d coordinate inputs and outputs the element-wise
seperable filter weight.
\vspace{-0.15in}

\paragraph{Implementation details:}
The backbone network consists of five residual blocks with [2, 4, 8, 12, 12] convolution layers with
channel dimensions [32, 64, 128, 192, 256] respectively. The initial convolution for each residual
block has stride 2 to downsample the feature map. The grid unit lengths are 0.15m for car,
0.5m/0.2m/0.1m/0.05m for pedestrian, and 0.33m for cyclist. We use 3$\times$3$\times$3 filters for
car and pedestrian, and 7$\times$7$\times$7 filters for cyclist. NMS thresholds are 0.1, 0.3, and
0.5 for car, pedestrian, cyclist respectively. For data augmentation we apply random scaling of
0.9$\times$ to 1.1$\times$ random rotation of -5 to 5 degrees along $z$-axis and random translation
of -5 to 5 meters for $x$ and $y$ axes. Models are trained using SGD with momentum for 50 epochs
with mini-batch size 16. The initial learning rate is 0.01 with 0.1$\times$ learning rate decay at
the 30th and 45th epoch; weight decay constant is 2e-4.
\vspace{-0.2in}

\paragraph{Results and discussion:}
In Table~\ref{tab:kitti_val} we show results on KITTI validation set where we compare our proposed
operator with the HDNet baseline and PCC \cite{pccn}. Overall, our method has the best performance
across all three categories, especially on pedestrian and cyclist. Notably, PCC has a negative
impact on the cyclist detector, possibly due to less training data and sparser point clouds for
positive examples, whereas our method delivers a significant gain over the baseline network. In
Figure~\ref{fig:filterviz}, we visualize the 3D filters learned on our cyclist detector, compared to
the ones learned by PCC. Our filters learn more expressive 3D shapes that are not linearly
separable.
\subsection{ModelNet 40 classification}
To further verify the effectiveness of our proposed operator, we evaluate on
ModelNet~\cite{shapenet}, a standard point cloud classification benchmark. ModelNet contains CAD
models of 40 categories, 9,843 shapes for training and 2,468 for testing.

\begin{table}
\begin{center}
\begin{small}
\begin{tabular}{|c|c|c|}
\hline
Method                      & \# points& Acc.                \\
\hline\hline                                                   
DeepSets \cite{deepsets}    & 1000     & 87.1                \\
ECC \cite{ecc}              & 1000     & 87.4                \\
PointNet \cite{pointnet}    & 1024     & 89.2                \\
FlexConv \cite{accv2018/Groh}    & 1024     & 90.2                \\
SpiderCNN \cite{spidercnn}  & 1024     & 90.5                \\
kd-net \cite{kdnet}         & 1024     & 90.6                \\
PointNet++ \cite{pointnet2} & 1024     & 90.7                \\
SO-Net \cite{li2018sonet}         & 2048     & 90.9                \\
MCConv \cite{mcconv}        & 1024     & 90.9                \\
PointCNN \cite{li2018pointcnn}    & 1024     & 92.2                \\
DGCNN    \cite{dgcnn}       & 1024     & 92.2                \\
KPConv   \cite{thomas2019KPConv} & ~6800 & \tb{92.9}          \\      
\hline
Ours Best                   & 1024     & 91.7           \\
\hline
\end{tabular}
\end{small}
\vspace{0.1in}
\caption{ModelNet-40: Classification results}
\label{tab:modelnet}
\end{center}
\end{table}

\begin{table}
\begin{center}
\begin{small}
\begin{tabular}{|c|c|c|}
\hline
Method                      & Acc.                \\
\hline\hline                                              
PointNet \cite{pointnet}    & 89.2                \\
+DeformFilters (Ours)       & +1.8                \\
\hline                                               
PointNet++ \cite{pointnet2} & 90.7                \\
+DeformFilters (Ours)       & +1.0               \\
\hline
\end{tabular}
\end{small}
\vspace{0.1in}
\caption{ModelNet-40: Gain in accuracy
}
\label{tab:modelnetabl}
\end{center}
\end{table}
\paragraph{Implementation details:} We modified the PointNet \cite{pointnet} and PointNet++
\cite{pointnet2} architectures to incorporate our deformable filter layers. The inputs to these
networks are $xyz$ coordinates (3 channels). In the original PointNet, we replace the pointwise
fully connected layers with deformable filter convolution layers. The resulting architecture has the
same number of channels ([3, 64, 64, 64, 128, 1024]). In PointNet++, since our convolution layer
operates on a different neighborhood compared to the sampling and grouping layers, we augment the
original architecture with a residual branch that contains the deformable filter layers with the
same number of hidden units compared to the MLP network in the original network. We use
3$\times$3$\times$3 convolution filters with filter grid unit length 0.2 for the PointNet
architecture, and [0.2, 0.4, 0.8] for each downsample stage of the PointNet++ architecture. We fix
the neighborhood size to be 8. We use mini-batch size 16, and base learning rate 1e-3 with an
exponential decay of 0.7$\times$ every 20 epochs. In both experiments, we use the standard 1,024
points with furthest point samples as inputs to the network, and for fair comparison with baselines
we do not aggregate the final prediction from multiple votes.
 \vspace{-0.2in}

\paragraph{Results and discussion:} As shown in Table~\ref{tab:modelnet} and \ref{tab:modelnetabl}, by
simply adding our deformable filter layers in standard PointNet-based architecture, we observe a
reasonable increase in performance, comparable to other competitive approaches using point cloud
inputs.


\section{Conclusion}
This paper presents deformable filter convolution, a learnable convolution layer that combines
voxel-like filtering and spatially continuous reasoning. It naturally augments existing
top-performing voxel-based network architectures by fusing point features into input and output
branches. We show significant gain in performance of the joint network, compared to voxel only
baselines and other point-based convolution approaches. The proposed convolution operator enables us
to achieve state-of-the-art performance on LiDAR semantic segmentation. As future work, we plan to
integrate our proposed convolution layer as a fundamental building block of  end-to-end
point-based networks, without resorting to voxelized feature maps.

\clearpage

{\small
\bibliographystyle{ieee}
\bibliography{ref}
}
\appendix
\newpage
\section{Technical proofs}
\label{sec:proofs}
\setlength\parindent{0pt}
\textbf{Proposition~\ref{translation}} (Translation equivariance).
\begin{proof}
Let $\tilde{\bx} = \bx - \Delta \bx$, we first show that the neighborhood can be translated:
\begin{align}
{\cal{N}}(\by + \Delta \bx)
=& \{\bx: \bx \in N(\by + \Delta \bx)\} \nonumber \\
=& \{\bx: \lVert \bx - \by - \Delta \bx \rVert  \le r\} \nonumber \\
=& \{\tilde{\bx} + \Delta \bx: \lVert \tilde{\bx} - \by \rVert  \le r\} \nonumber \\
=& \{\tilde{\bx} + \Delta \bx: \tilde{\bx} \in {\cal{N}}(\by)\}.
\end{align}
For all $\by \in \mathbb{R}^d$, 
\begin{align}
&\mathcal{T}^H_{\Delta \bx}(C_{g}(f))(\by) =h(\by + \Delta \bx) \nonumber \\
&=
\sum_{\bx \in N(\by + \Delta \bx)} f(\bx) \cdot \left[ \sum_{\bx' \in X'} k(\bx', \by + \Delta \bx - \bx) g(\bx') \right]  \nonumber\\
&=
\sum_{\tilde{\bx} \in N(\by)} f(\tilde{\bx} + \Delta \bx) \cdot \left[ \sum_{\bx' \in X'} k(\bx', \by - \tilde{\bx}) g(\bx') \right]  \nonumber\\
&=
\sum_{\tilde{\bx} \in N(\by))} \mathcal{T}^F_{\Delta \bx}(f)(\tilde{\bx}) \cdot \left[ \sum_{\bx' \in X'} k(\bx', \by - \tilde{\bx}) g(\bx') \right] \nonumber\\
&=C_{g}(\mathcal{T}^F_{\Delta \bx}(f))(\by).
\end{align}
\end{proof}

\setlength\parindent{0pt}
\textbf{Proposition~\ref{permutation}} (Permutation equivariance).
\begin{proof}
Let $s$ be any element in $\text{Sym}(M)$. Since $s$ is bijective, let $\tilde{j} = s^{-1}(j),
s(N(i)) \equiv \{s(j): j \in N(i)\}$, $\tilde{N} = s^{-1}(N(s(i)))$. For all $i \in \mathbb{Z}_M$,
\begin{align}
&\mathcal{P}^H_s(C_{g}(p,f))(i) = h(s(i)) \nonumber \\
&=
\sum_{j \in N(s(i))} f(j) \cdot \left[ \sum_{\bx' \in X'} k(\bx', p(s(i)) - p(j)) g(\bx') \right]  \nonumber\\
&=
\sum_{\tilde{j} \in \tilde{N}} f(s(\tilde{j})) \cdot
\left[ \sum_{\bx' \in X'} k(\bx', p(s(i)) - p(s(\tilde{j})) g(\bx') \right]  \nonumber\\
&=
\sum_{\tilde{j} \in \tilde{N}} \mathcal{P}_s^F(f)(\tilde{j}) \cdot
\left[ \sum_{\bx' \in X'} k(\bx', \mathcal{P}_s^P(p)(i) - \mathcal{P}_s^P(p)(\tilde{j}) g(\bx') \right]  \nonumber\\
&=C_{g}(\mathcal{P}^{P \times F}_{s}(p,f))(i).
\end{align}
\end{proof}

\section{KITTI detection results}
\begin{figure*}[h]
\centering
\begin{minipage}{\textwidth}
\centering
\footnotesize
\begin{tabular}{c c c}
HDNet & +PCC & +DeformFilter \\
\includegraphics[width=0.3\textwidth,trim={0 4cm 0 4cm},clip]{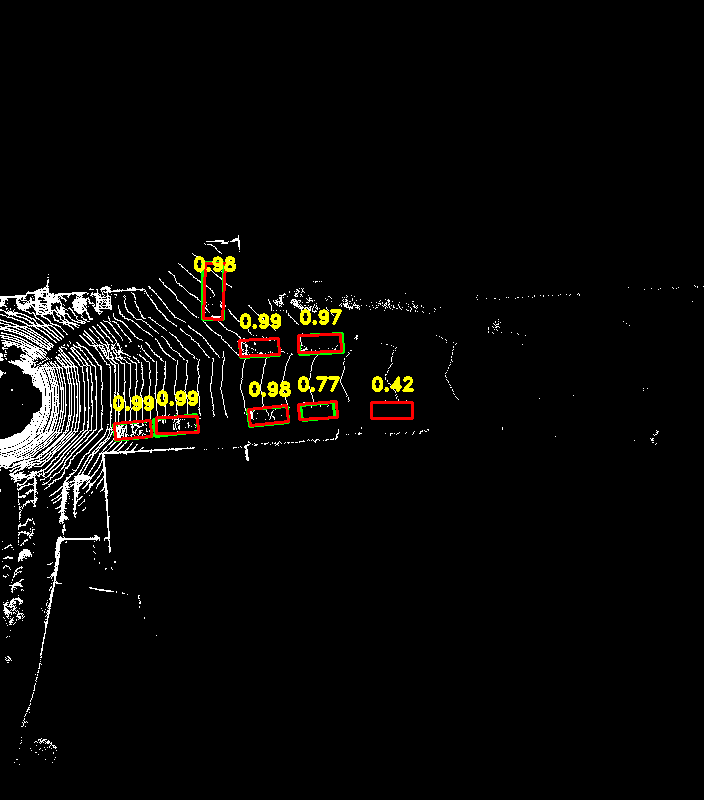} &
\includegraphics[width=0.3\textwidth,trim={0 4cm 0 4cm},clip]{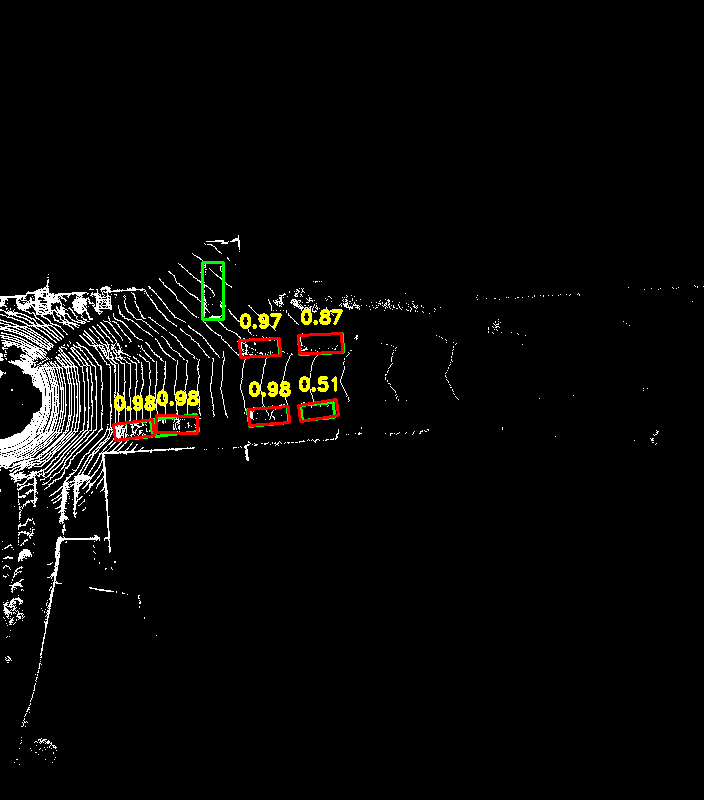} &
\includegraphics[width=0.3\textwidth,trim={0 4cm 0 4cm},clip]{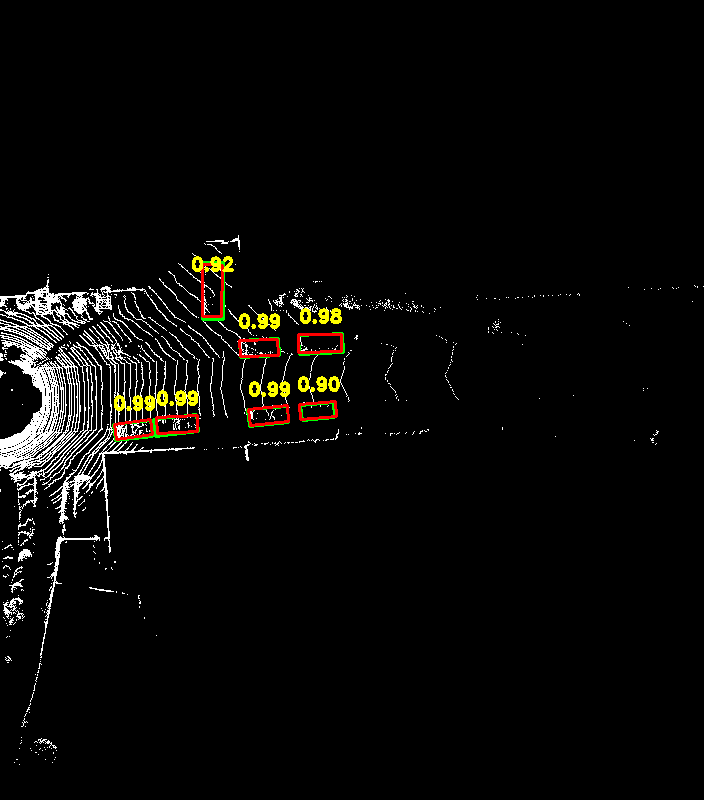} \\
\\
\hline
\\
\includegraphics[width=0.3\textwidth,trim={0 4cm 0 4cm},clip]{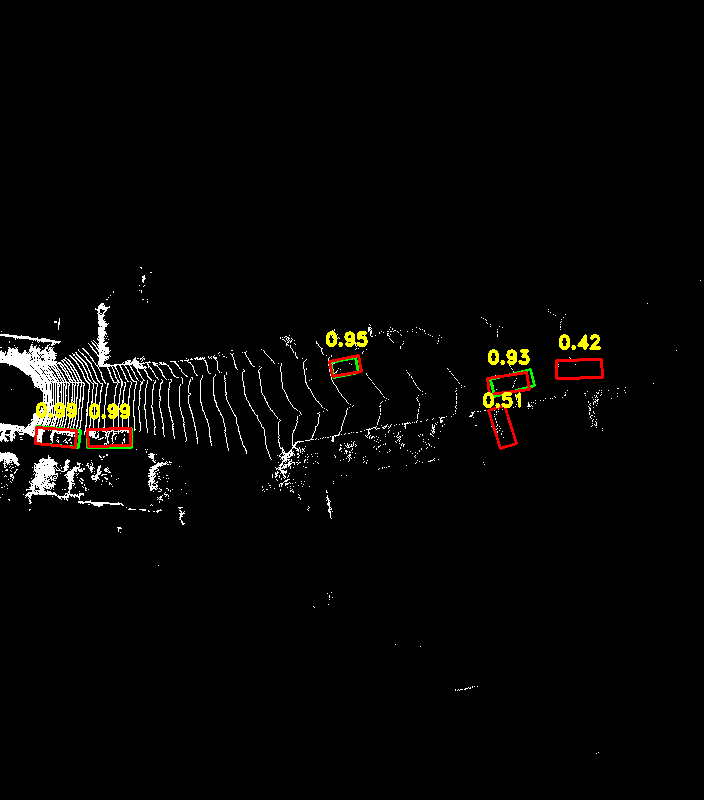} &
\includegraphics[width=0.3\textwidth,trim={0 4cm 0 4cm},clip]{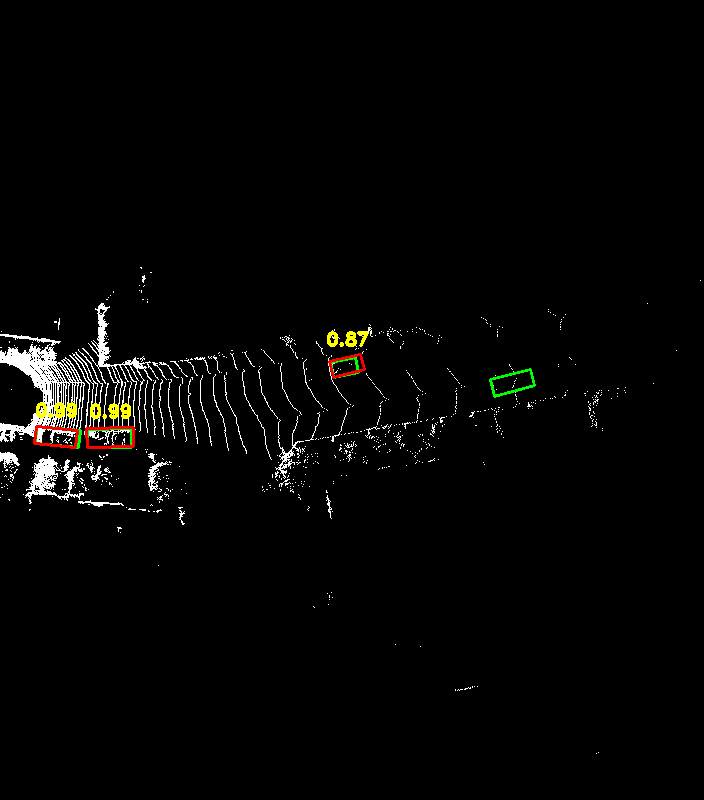} &
\includegraphics[width=0.3\textwidth,trim={0 4cm 0 4cm},clip]{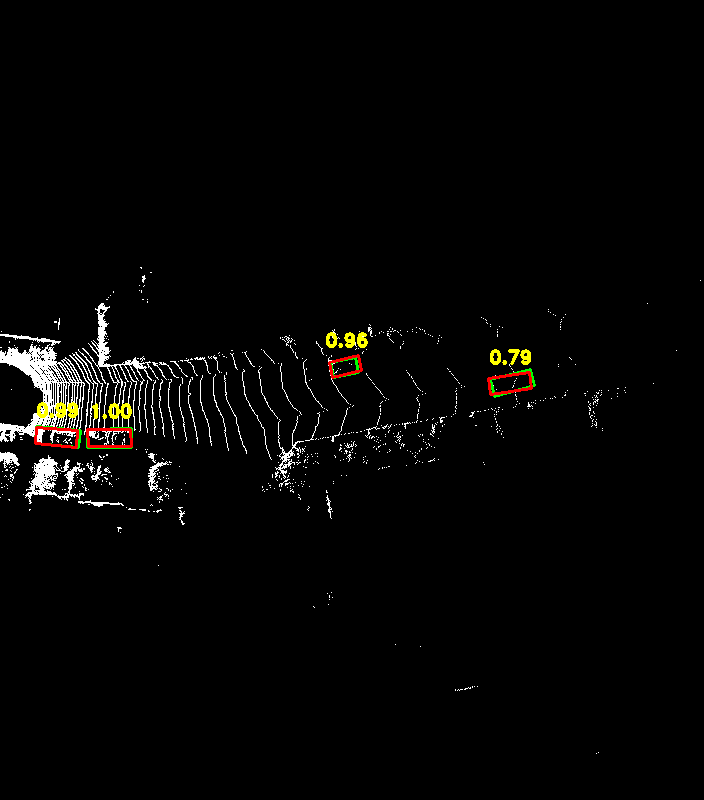} \\
\\
\hline
\\
\includegraphics[width=0.3\textwidth,trim={0 4cm 0 4cm},clip]{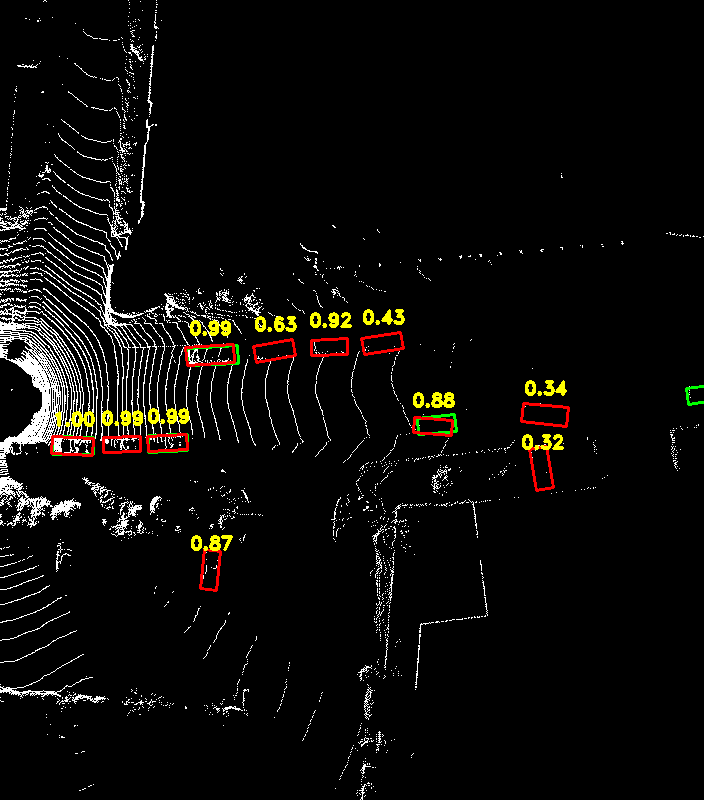} &
\includegraphics[width=0.3\textwidth,trim={0 4cm 0 4cm},clip]{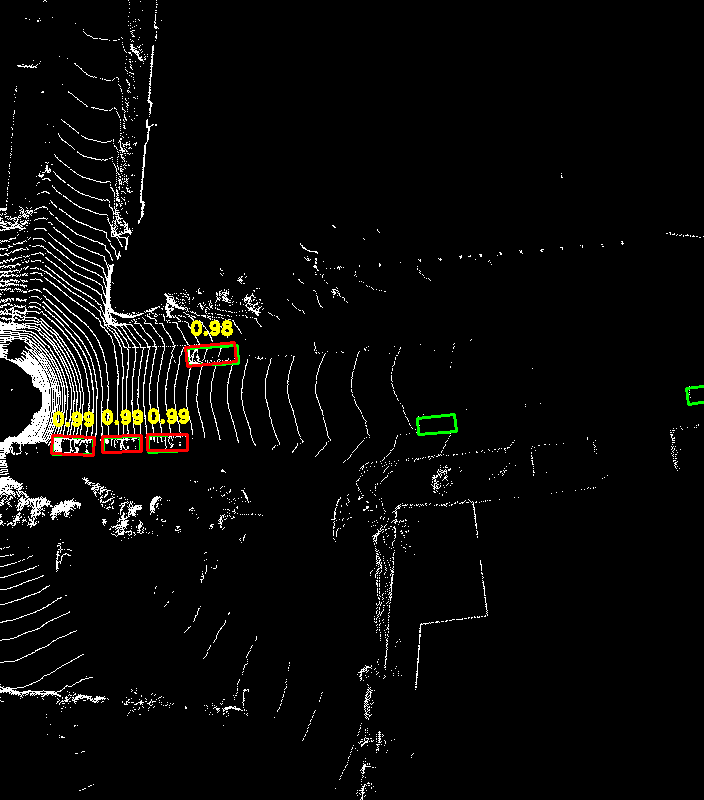} &
\includegraphics[width=0.3\textwidth,trim={0 4cm 0 4cm},clip]{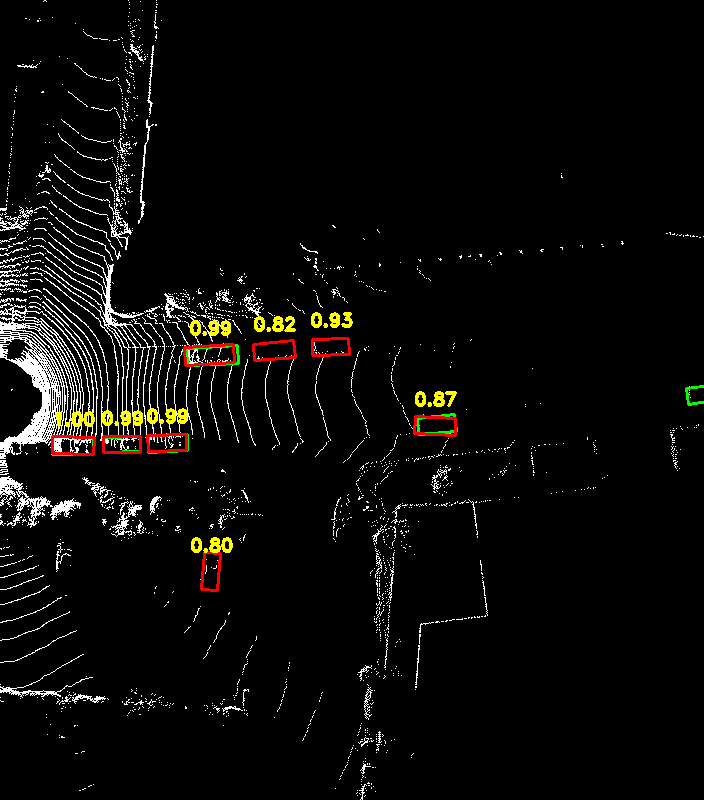} \\
\\
\hline
\\
\includegraphics[width=0.3\textwidth,trim={0 4cm 0 4cm},clip]{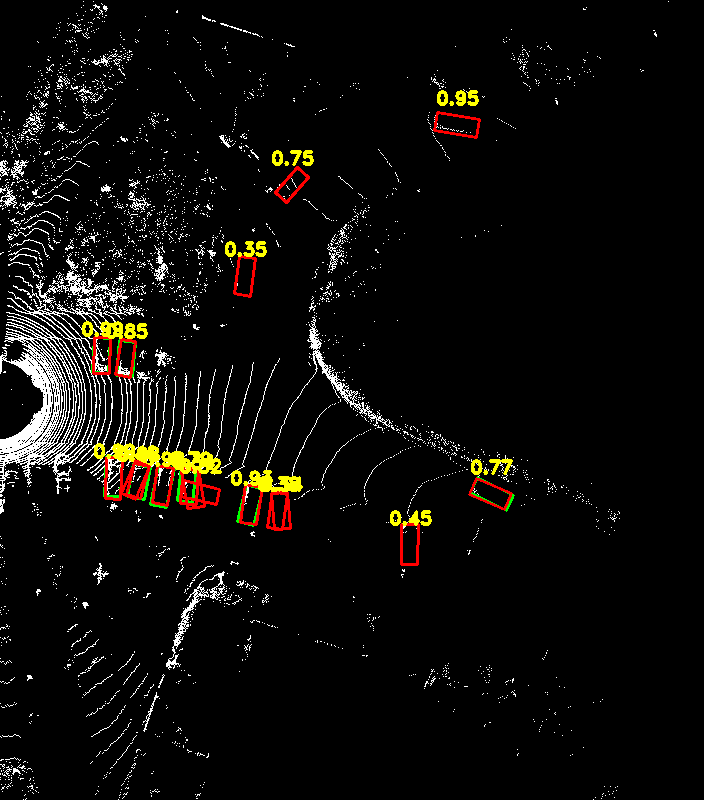} &
\includegraphics[width=0.3\textwidth,trim={0 4cm 0 4cm},clip]{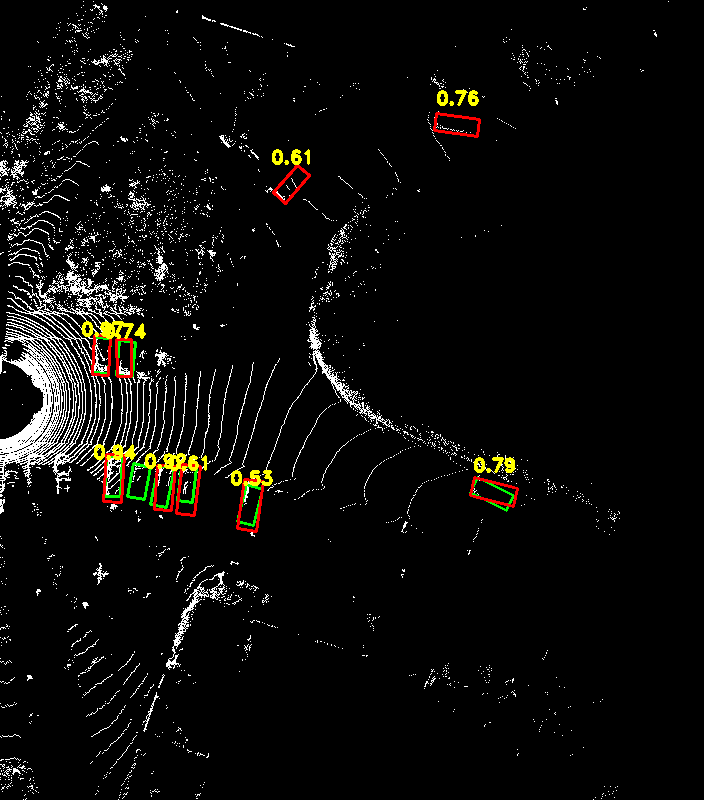} &
\includegraphics[width=0.3\textwidth,trim={0 4cm 0 4cm},clip]{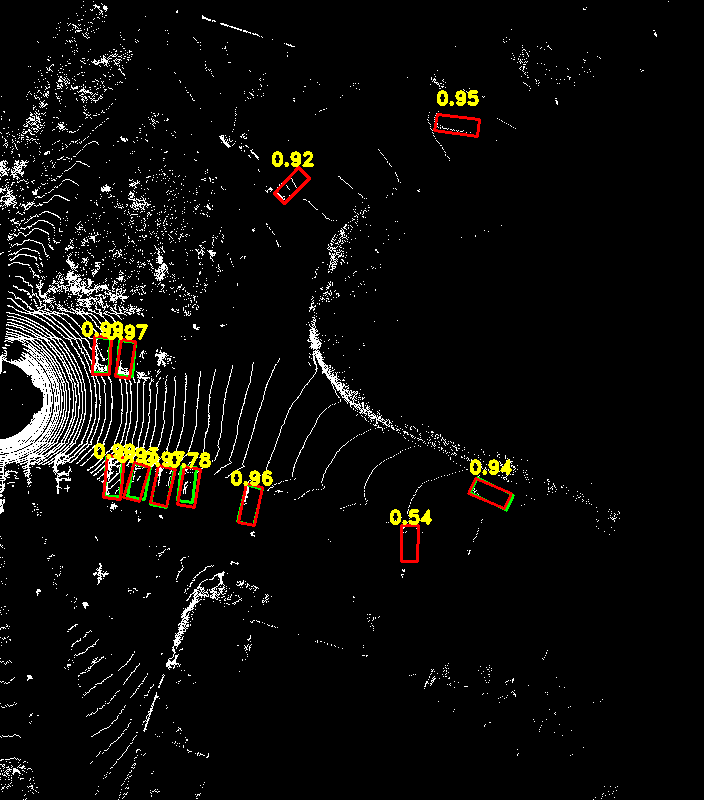} \\
\end{tabular}
\end{minipage}
\vspace{0.05in}
\caption{Visualization of results on KITTI BEV LiDAR detection on cars}
\label{fig:kittidet}
\end{figure*}
Figure~\ref{fig:kittidet} shows vehicle detection results on the KITTI dataset.

\end{document}